\documentclass[11pt]{article}

\usepackage[final]{acl}

\usepackage{times}
\usepackage{multirow}
\usepackage{latexsym}
\usepackage[utf8]{inputenc}
\usepackage{booktabs}  
\usepackage[table]{xcolor} 
\usepackage{caption}
\usepackage{pifont}
\usepackage{amssymb} 

\definecolor{headerblue}{HTML}{BEDBFF} 
\definecolor{grouporange}{HTML}{EFF6FF} 
\definecolor{cmarkgreen}{HTML}{32CB00} 
\definecolor{xmarkred}{HTML}{FF0000}   
\definecolor{deepred}{HTML}{BB0C0B}   
\newcommand{\up}[1]{\textcolor{red}{\textbf{\,\(\uparrow\)#1}}}

\newcommand{\xmark}{\textcolor{xmarkred}{\ding{55}}}
\newcommand{\cmark}{\textcolor{cmarkgreen}{\ding{51}}}

\usepackage[T1]{fontenc}

\usepackage[utf8]{inputenc}

\usepackage{microtype}

\usepackage{inconsolata}

\usepackage{graphicx}

\title{Med-R$^{2}$: An Adversarial Benchmark \\ for Evidence-Grounded Reasoning in Medical VLMs}

\author{
 \textbf{Wen Ma\textsuperscript{1}},
 \textbf{Fucheng Niu\textsuperscript{1}},
 \textbf{Zhiting Fan\textsuperscript{1}},
 \textbf{Zikai Xiao\textsuperscript{1}},
\\
 \textbf{Jiaxiang Liu\textsuperscript{3}},
 \textbf{Zuozhu Liu\textsuperscript{1}}
\\
\\
 \textsuperscript{1}Zhejiang University, Zhejiang, China,
 \textsuperscript{3}Guangdong Institute of Intelligence Science and Technology
\\
}

\begin{document}
\maketitle
\begin{abstract}

Vision–language models (VLMs) have demonstrated impressive capabilities in general medical visual question answering, yet due to limited interpretability, it remains unclear whether their predictions reflect evidence-grounded clinical reasoning or reliance on spurious priors. We introduce \textbf{Med-R$^{2}$ Bench}, a hierarchical benchmark aligned with the clinical workflow to evaluate adversarial robustness with visual grounding.
We design stepwise QA tasks to assess whether reasoning chains are strictly grounded in visual evidence across the four clinical stages, and employ adversarial perturbations to test robustness against misleading cues.
Med-R$^{2}$ comprises 42,432 images, 31 task categories, and 110,406 QA pairs. Evaluation across 14 VLMs reveals a sequential performance degradation along the four-stage clinical workflow.
Adversarial experiments show that models rely heavily on correct prompts to guess answers. Even when provided with explicit visual cues, the models struggle to accurately align textual descriptions.
Finally, we demonstrate stepwise fine-tuning using our hierarchical data significantly improves reasoning robustness, highlighting its potential to drive future improvements in evidence-based medical AI.

\end{abstract}

\section{Introduction}
AI-driven diagnostic systems face a major barrier to clinical deployment: their decision-making processes often lack interpretability, hindering clinicians' ability to trust and verify predictions in high-stakes settings\cite{87,36}. 
Recently,  VLMs have emerged as a promising approach to explainable medical AI, as they can provide natural-language rationales\cite{29,18,14,27,74,35,90,liu2024medcot}.
Such models could articulate a clinician-like reasoning process, linking visual evidence with clinical semantics\cite{72,13,66,wang2025v2t}. In clinical workflows, physicians begin with visual cues, localize relevant anatomy, characterize lesions, and integrate evidence to form a diagnosis.

However, existing evaluation methods mostly focus on the accuracy of the final answer, making it difficult to reveal the vulnerabilities exposed by the model when encountering misleading questions or adversarial visual disturbances. At the same time, there is limited insight into whether the model's reasoning trajectory follows the clinical evidence chain, overlooking whether the model truly performs correct process reasoning based on medical images\cite{cai,gai2025medthink}.

\begin{figure}[t]
\centering
\vspace{-4mm}
\includegraphics[width=0.48\textwidth]{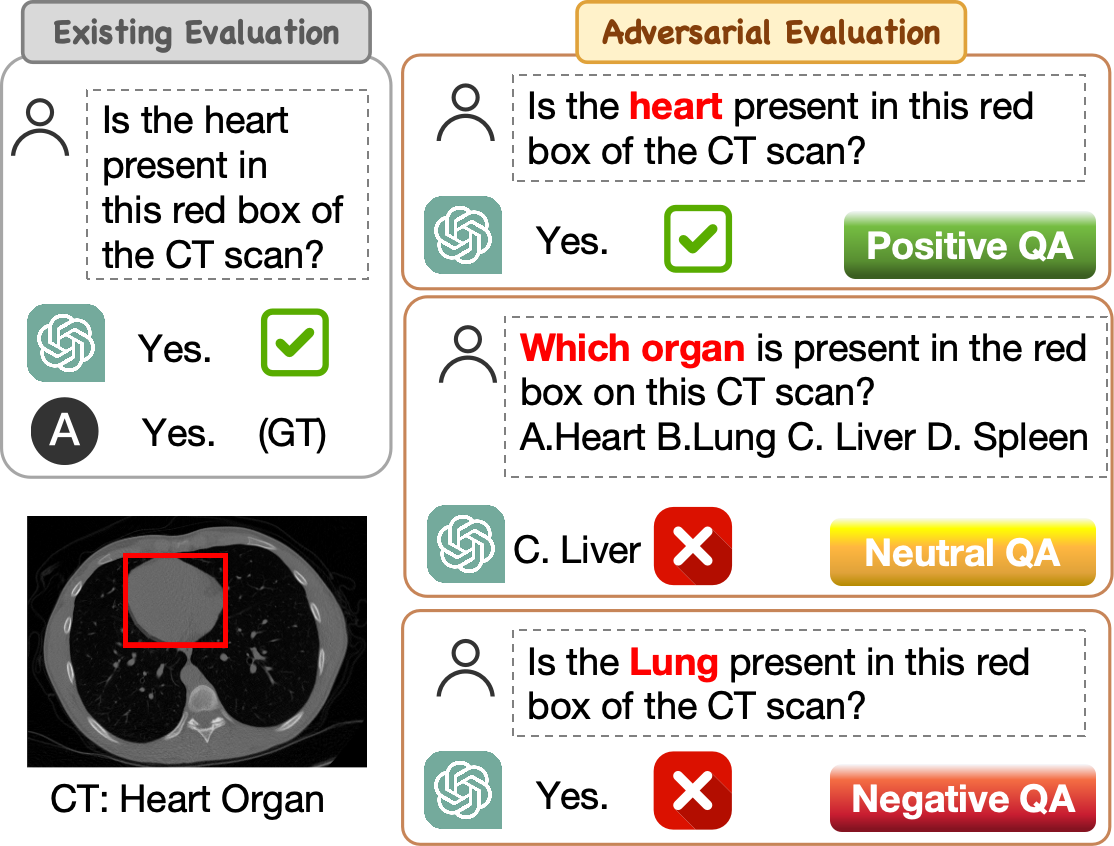} 
\vspace{-5mm}

\caption {Comparison of standard (left) versus adversarial evaluation (right) for medical visual question answering. The adversarial framework (right) uses designed positive, neutral, and negative QA to test if a model's reasoning is robust.}
\label{sample_eg}
\vspace{-5mm}
\end{figure}

\begin{figure*}[t]
\centering
\vspace{-5mm}
\includegraphics[width=1\textwidth]{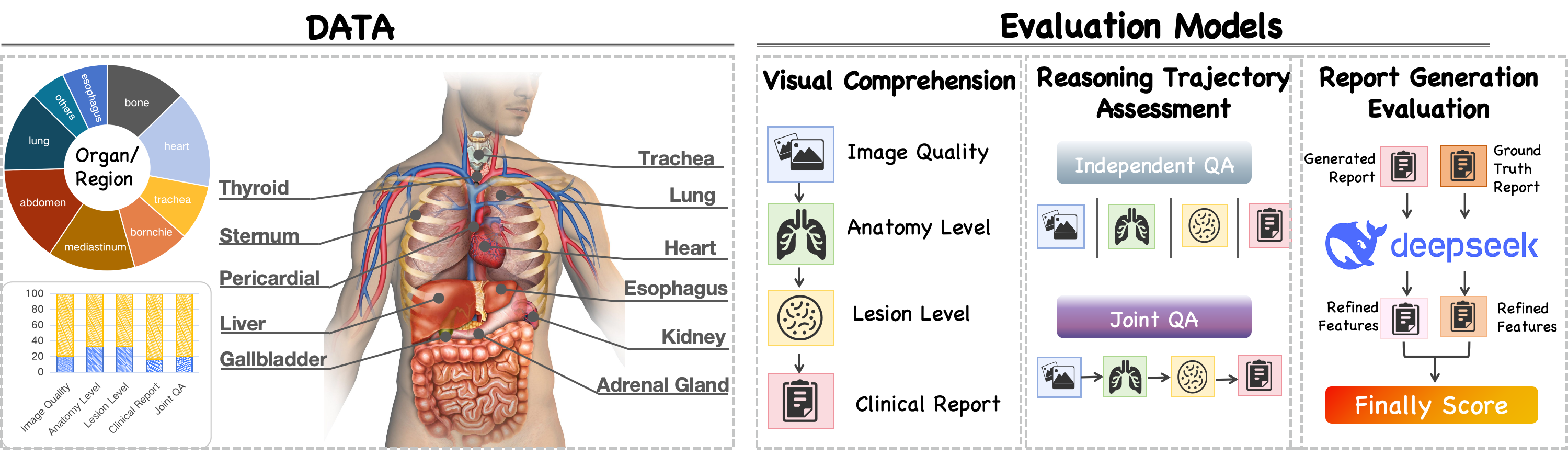} 
\vspace{-5mm}
\caption { Overview of the Med-R$^{2}$ Bench, where R$^{2}$ denotes Evidence-grounded Reasoning and Robustness.}
\label{Fig1}
\vspace{-5mm}
\end{figure*}

To address this gap, we introduce Med-R$^{2}$ that evaluates two critical capabilities of medical VLMs: their ability to form an evidence-driven reasoning chain and the reasoning robustness. To assess the reasoning chain, Med-R² employs a four-level hierarchical structure aligned with the clinical workflow. We utilize independent QA to test the extraction of specific visual evidence and joint QA to evaluate coherent, multi-step reasoning to ensure a model’s logic is explicitly tied to visual cues at each intermediate stage. To assess robustness, we introduce adversarial samples with crafted distractors, which serve as a stress test to determine if a model's reasoning can resist misleading information. Our dual evaluation strategy allows us to reliably distinguish models that genuinely comprehend medical images from those that simply rely on superficial pattern matching.

We conducted a comprehensive evaluation of 14 publicly available VLMs, encompassing both general-purpose and medical-specialized models (see Table~\ref{tab:compare_models}). 
Our key findings are as follows: 

(I)As task complexity increases along the clinical reasoning hierarchy—from \textbf{image quality} to \textbf{anatomy level}, \textbf{lesion level}, and ultimately \textbf{clinical report generation}—model performance consistently deteriorates, with lesion-level reasoning emerging as the primary bottleneck. This underscores the gap between visual recognition and evidence-based clinical inference. 

(II)The adversarial evaluation reveals a significant decline in the model's performance with the introduction of adversarial interference.
Robustness follows a clear progression (\textbf{Positive $>$ Neutral $>$ Negative}). The model tends to follow prompts in a way that under Negative, it abandons visual evidence; under Positive, scores become inflated. This suggests that it relies more on statistically correlated "answer-patterns" than on closed-loop verification grounded in evidence.

(III)Despite high-quality grounded annotations, models exhibit limited localization accuracy and lack fine-grained structural understanding, indicating that current representations fall short of achieving robust medical vision-language alignment and spatial correspondence.

To further validate the effectiveness of our benchmark, we evaluated the impact of fine-tuning with Hulu-Med.
Fine-tuning resulted in significant improvements across all metrics, especially in handling high-difficulty tasks and negative samples, demonstrating enhanced robustness. This highlights the crucial role of high-quality data in training robust models.

\begin{table*}[t]
\centering
\vspace{-5mm}
\caption{Comparison of Medical VLM Benchmarks}
\vspace{-3mm}
\label{tab:compare_bench}

\renewcommand{\arraystretch}{1.3} 
\setlength{\tabcolsep}{1pt}       

\small 

\begin{tabular}{l@{\hspace{0.25cm}}lccccc}
\toprule
\rowcolor{headerblue} 
\textbf{Benchmark} 
& \textbf{Imaging Modalities} 
& \textbf{Dataset Scale}
& \textbf{Grounded}
& \textbf{Reasoning} 
& \textbf{Adversarial} 
& \textbf{Task Types} \\ \midrule

VQA-RAD & Radiography, CT &315 images; 3,515 QA & \xmark & \xmark  & \xmark & VQA \\

SLAKE & Radiography & 642 images; 14,028 QA & \cmark & \xmark  & \xmark & VQA \\

VQA-Med & CT, MRI & 3200 images; 12,792 QA & \xmark & \xmark   & \xmark & VQA \\

PMC-VQA & CT, MRI, others &149k images; 227k QA & \xmark & \xmark  & \xmark & VQA \\

Rad-ReStruct & Radiography & 3,720 images; 3,597 QA & \xmark & \xmark  & \xmark & VQA \\

PathMMU & Pathology & 24,067 images; 33,428 QA& \xmark & \xmark  & \xmark & VQA \\

DrVD-Bench & 5 modalities & 7,789 images; 8276 QA & \xmark & \cmark  & \xmark & VQA, MRG \\

MultiMedEval & 11 modalities & 133,521 images; 68,720 QA & \xmark & \xmark  & \xmark & VQA, Open QA, Others \\

OmniMedVQA & 12 modalities &  118,010 images; 127,995 QA & \xmark & \xmark  & \xmark & VQA \\

\rowcolor{grouporange} 
\textbf{Med-R$^{2}$ Bench} & CT, MRI & 42,432 images; 110,406 QA & \cmark & \cmark & \cmark & VQA,Open QA, MRG  \\ 

\bottomrule
\vspace{-9mm}
\end{tabular}
\end{table*}

\section{Related Work}

Vision-Language Models (VLMs) have evolved from early feature alignment to end-to-end reasoning. Technical foundations were laid by BERT~\cite{koroteev2021bert} and ViT~\cite{dosovitskiy2020image}, followed by CLIP~\cite{radford2021learning}, which achieved robust visual-textual alignment. Subsequently, LLaVA~\cite{liu2023visual} introduced general conversational capabilities through instruction tuning, while models like GPT-4o~\cite{hurst2024gpt}, Qwen2.5-VL~\cite{bai2025qwen2}, and Gemini 2.5~\cite{comanici2025gemini} have pushed zero-shot generalization to new heights. In the medical domain, specialized models have made continuous breakthroughs via domain-specific fine-tuning: LLaVA-Med~\cite{LLAVA-med} leveraged biomedical literature to enhance professional perception, Med-Flamingo~\cite{Med-Flamingo} strengthened clinical few-shot learning, and Med-Gemini~\cite{Med-Gemini} integrated long-context reasoning with clinical search capabilities. However, despite their impressive performance, ensuring their reasoning processes maintain consistency with clinical evidence remains a critical challenge in current research.

Medical Visual Question Answering (Med-VQA) benchmarks are evolving from single-task evaluations toward multi-dimensional clinical reasoning. Early benchmarks like VQA-RAD~\cite{VQARAD} and SLAKE~\cite{SLAKE} established the foundations for radiology QA and knowledge-enhanced reasoning. Subsequently, Rad-ReStruct~\cite{Rad-restruct} and PathMMU~\cite{sun2024pathmmu} introduced hierarchical report generation and expert-level pathology reasoning into the evaluation framework. To reflect the
clinical reasoning workflow , DrVD-Bench~\cite{DrVD} proposed reasoning trajectory assessment to verify clinical logical consistency. Meanwhile, OmniMedVQA~\cite{hu2024omnimedvqa} challenges the generalization boundaries of models through large-scale, multi-modal data covering diverse anatomical regions. Finally, toolkits such as MultiMedEval~\cite{royer2024multimedeval} have standardized evaluation protocols, providing a unified pathway for the fair assessment of medical VLMs.

\section{Design of Med-R$^{2}$ Bench}
\subsection{Overview}

We introduce Med-R$^{2}$ Bench, a multi-scale evaluation for medical VLMs. Grounded in precisely annotated image-text pairs, the benchmark ensures that tasks are tied to real-world data, enabling realistic and robust assessments, especially for medical text generation and understanding.

\begin{figure*}[h!]
\centering
\vspace{-5mm}
\includegraphics[width=1\textwidth]{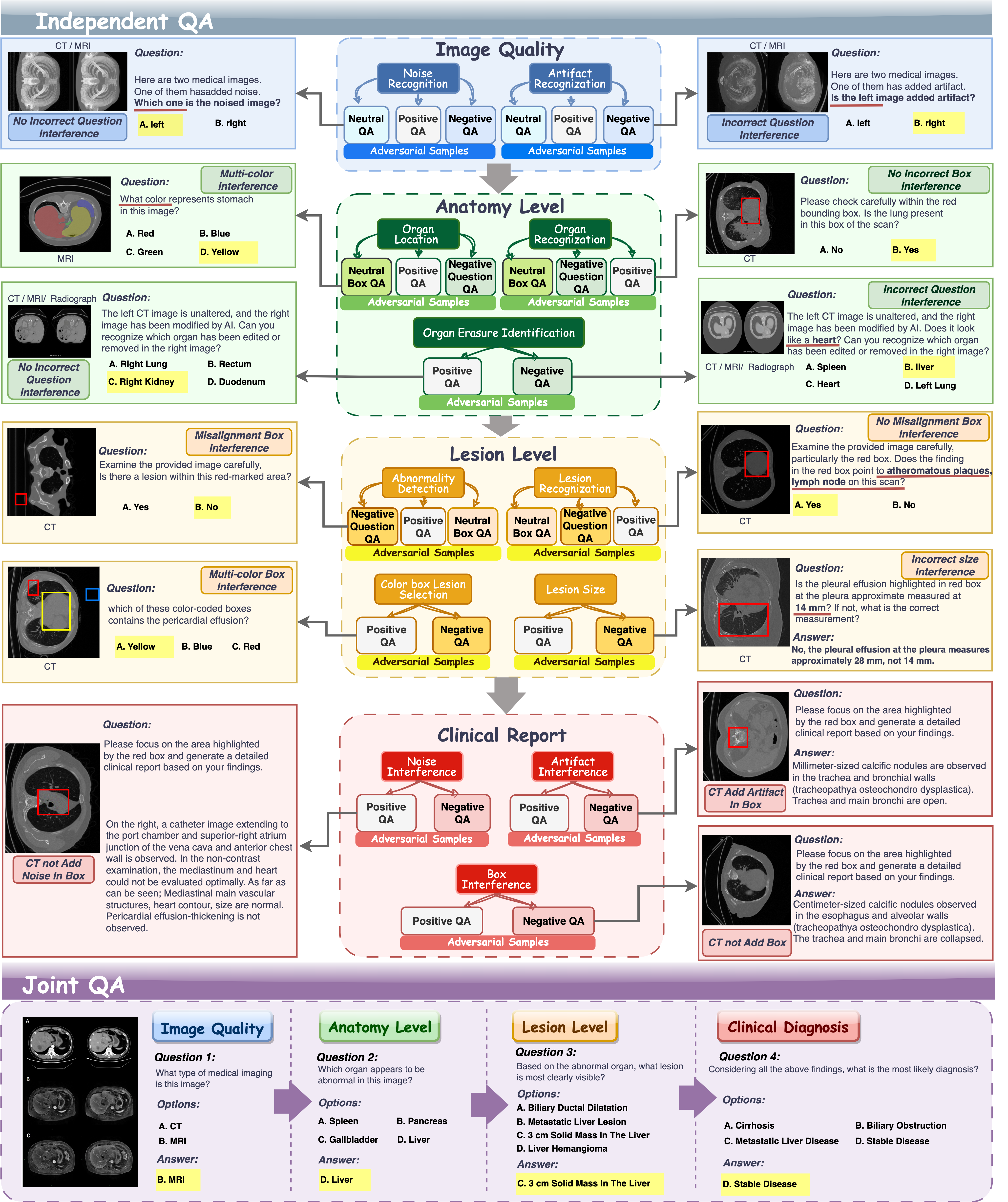} 
\vspace{-7mm}
\caption{Med-R$^{2}$: a hierarchical evaluation framework for evidence-based reasoning in medical imaging. Representative tasks are organized along a clinical reasoning cascade, and Positive/Neutral/Negative adversarial variants are introduced to assess robustness under misleading cues. Joint QA setting further chains multi-step questions into an end-to-end reasoning path to evaluate cross-level coherence and consistency.}

\label{pipeline}
\vspace{-6mm}
\end{figure*}

Med-R$^{2}$ Bench evaluates VLMs from three key perspectives: (i) the reliability of visual evidence grounding, (ii) the robustness of clinical reasoning under adversarial conditions, and (iii) the ability to generate accurate medical reports. 
As shown in Fig.~\ref{pipeline}, the framework mirrors clinical diagnostic workflows, allowing for a more comprehensive evaluation of model performance in complex, real-world tasks. This approach tests models on both visual grounding and clinical reasoning in adversarial settings, positioning Med-R$^{2}$ Bench as a critical tool for advancing medical AI systems.
The benchmark includes three complementary modules that assess grounded visual understanding and clinical reasoning under adversarial conditions:

\textbf{Visual Comprehension} consists of 110,406 image-question pairs across 29 clinical tasks, structured by clinical reasoning depth, from image quality to lesion localization and diagnostic inference, with an explicit emphasis on grounding predictions in clinically relevant visual evidence.

\textbf{Adversarial Reasoning Assessment} includes 42,432 QA rounds, testing models' ability to perform stepwise clinical reasoning under joint and independent QA paradigms. Adversarial perturbations in visual and textual inputs assess reasoning robustness, consistency, and reliance on evidence.

\textbf{Open-ended Evaluation} tests free-form report generation and open-ended medical question answering, evaluating whether models maintain grounded understanding in unconstrained outputs.

\begin{table}[ht!]
    \centering
    \vspace{-1mm}
    \caption{Statistics of Med-R$^{2}$ Bench}
    \vspace{-3mm}
    \resizebox{0.9\linewidth}{!}{
    \small
    \begin{tabular}{lrr}
        \toprule[1.5pt] 
        
        \rowcolor{headerblue}
        \textbf{Category} & \textbf{Metric} & \textbf{Count} \\
        \toprule[1.5pt] 
        
        \rowcolor{grouporange}
        \multicolumn{3}{l}{\textit{\textbf{Module 1 Visual Comprehension}}} \\
        Total QA pairs & QA pairs & 110,406 \\
        Image Quality & QA pairs & 8764 \\
        Anatomy Level& QA pairs & 21,572 \\
        Lesion Level& QA pairs & 37,064 \\
        Clinical Report & QA pairs & 40,818 \\
        \midrule 
        \rowcolor{grouporange}
        \multicolumn{3}{l}{\textit{\textbf{Module 2 Inference Evaluation Module}}} \\
        Independent QA & QA pairs & 61,876 \\
        Joint QA & QA pairs & 3,088 \\
       
        \midrule 
        \rowcolor{grouporange}
        \multicolumn{3}{l}{\textit{\textbf{Module 3 Open-ended Evaluation}}} \\
        Open-ended QA & QA pairs & 1,536 \\
        Report Generation   & &42,354 \\
        \midrule 
   
        \rowcolor{grouporange}
        \multicolumn{3}{l}{\textit{\textbf{Global Dataset Statistics}}} \\
        Total images &images & 42,432 \\
        CT & images & 36,364 \\
        MRI & images & 6,068 \\
        Tasks  &classes & 31 \\
        Organ/Tissue  &classes & 193 \\
        Lesion  &classes & 212 \\
        Diagnosis Categories & classes & 170 \\
       
        \bottomrule[1.5pt] 
    \end{tabular}
    }
    \label{tab:statistics}
    \vspace{-5mm}
\end{table}

\subsection{Task Construction}

To systematically assess models’ foundational visual understanding ability, we design \textbf{Module 1: Visual Comprehension}, as shown in Table~\ref{tab:statistics}. It evaluates performance across four reasoning levels aligned with the clinical diagnostic pipeline:

\textbf{Image Quality}: This level focuses on simulating noise and artifacts in clinical settings, ensuring that medical images reflect potential disruptions encountered in actual diagnoses. To evaluate robustness, we introduce noise at three PSNR levels (20 dB, 25 dB, 35 dB)~\cite{DrVD} to simulate varying degrees of interference encountered in clinical scenarios, representing severe, moderate, and mild disruption. These settings mimic real-world conditions and assess model performance under different noise levels. In addition, we simulate motion artifacts, causing blurring and ghosting, which are common issues arising from patient movement during image acquisition. (See Fig.~\ref{L1-1},~\ref{L1-2})

\textbf{Anatomy Level}: In addition to recognizing and localizing anatomical organs and tissues in medical images, we also designed a color recognition task to assess the model's performance in complex scenarios by introducing interference through image color variations. The goal of this task is to evaluate whether the model can accurately associate visual cues with specific anatomical regions, especially when the image contains multiple colors that may affect image interpretation. (See Fig.~\ref{L2-1}-~\ref{L2-3})

\textbf{Lesion Level}: This level focuses on lesion recognition, localization, and morphology description. We use a pixel-level grounded localization approach for accurate identification and localization of anatomical and pathological regions, providing high precision in lesion and organ delineation. All annotations are grounded, ensuring a reliable foundation for diagnostic tasks. Additionally, we introduce a Color Box Lesion Selection task, where the model selects the box corresponding to a lesion, testing its ability to localize lesions accurately in complex scenarios. (See Fig.~\ref{L4-1}-~\ref{L4-4})
\textbf{Open-ended  Evaluation}:This level focuses on diagnostic classification and open-ended lesion size QA, In the report generation evaluation, the task is designed as a medical image annotation process, utilizing real-world references from the CT-RATE~\cite{28} clinic report text and RadGenome-ChestCT datasets~\cite{1}. These datasets offer detailed clinical interpretations (see Table~\ref{tab:datasets_modality}), ensuring that the generated reports are not only accurate in terms of diagnostic classification but also reflect a comprehensive clinical context. (See Fig.~\ref{L5-1}-~\ref{L5-3})

To evaluate models' reasoning capability, we introduce \textbf{Module 2: Inference Evaluation}.

but not only focuses on multi-level reasoning but also includes independent visual understanding. While the Visual Comprehension Module targets a single visual aspect per image, the Inference Evaluation Module integrates multiple reasoning levels—image quality, organ, lesion, and diagnosis—for the same image. Performance is evaluated using two prompt formats: Independent QA, where questions at different levels are assessed separately, and Joint QA, where all sub-questions are presented in a single prompt.

To evaluate holistic clinical interpretation and free-form medical language generation, we introduce \textbf{Module 3: Open-ended  Evaluation Evaluation} as the final stage. Built on 42,354 open-ended QA pairs, this module targets report-level reasoning. Given a medical image, models generate a clinically grounded report covering key findings, lesion location/morphology, relevant anatomy, and diagnostic implications. We also include open-ended QA (e.g., lesion size, extent, severity) without predefined options, requiring quantitative or semi-quantitative free-form responses to assess report writing and complex clinical querying.

\textbf{Adversarial Sample Design:}
To further enhance the robustness and diagnostic value of Med-R$^{2}$, we design adversarial samples and apply them consistently across all reasoning tasks. Each task includes two to three types of QA forms, with an equal number of Positive, Negative, and Neutral QAs to ensure a balanced evaluation.

\textbf{(i) Positive QA:} Aims to guide the model toward the correct reasoning path by providing clear and helpful prompts. These questions are designed to offer direct support, ensuring that VLMs can make accurate inferences and reach the correct answer through appropriate cues and bounding boxes.

\textbf{(ii) Negative QA:} Intentionally introduces misleading or distracting information in the questions to guide the model down incorrect reasoning paths. These prompts are designed to test the model's ability to resist interference and evaluate whether it truly understands the image content, rather than relying on memorization or overfitting to achieve high scores.

\textbf{(iii) Neutral QA:} Provides ambiguous or uncertain cues to assess the model's reasoning ability when no clear guidance is given. The goal is to evaluate how well the model can handle situations without any biased cues, testing robustness in making sound inferences without external direction.

Adversarial samples for each task are designed to ensure an equal distribution of the three QA types, enabling balanced assessment. The construction of these samples is shown in Fig.~\ref{pipeline}. More detailed QA settings, see Fig.~\ref{L1-1}-~\ref{L5-3}. By incorporating adversarial supervision, Med-R$^{2}$ provides a more rigorous evaluation of model reliability, error sensitivity, and adaptability in clinical scenarios.

\begin{table*}[h]
    \centering
    \vspace{-7mm}
    \small 
    \caption{Average results of different VLMs on Med-R$^{2}$: ACC (\%) is reported for the four-stage tasks and Joint QA, with the overall average; clinical reports are evaluated by ROUGE-1/BERTScore.}
    \vspace{-3mm}
    \renewcommand{\arraystretch}{1.2} 
    \resizebox{\linewidth}{!}{%

    \begin{tabular}{lccccc|c} 
        \toprule[1.5pt] 
        \rowcolor{headerblue}
        \textbf{Methods} 
        & \textbf{Image Quality} 
        & \textbf{Anatomy Level} 
        & \textbf{Lesion Level} 
        & \textbf{Joint QA} 
        & \textbf{Overall (Avg)} 
        & \textbf{Clinical Report} \\ 
        \toprule[1.5pt]
        
        \rowcolor{grouporange}
         \multicolumn{7}{c}{\textbf{Medical Specific}} \\ 
        Hulu-Med-4B & 41.4 & 60.3 & 57.9 & 58.4 & 54.5 & 13.7/83.3 \\ 
        Hulu-Med-7B & 36.5 & 60.7 & {\textbf{\textcolor{blue}{58.8}}} & 48.9 & 51.2 & {\textbf{\textcolor{red}{14.6}}}/{\textbf{\textcolor{blue}{83.6}}} \\
        Hulu-Med-14B & 47.4 & {\textbf{\textcolor{red}{68.1}}} & {\textbf{\textcolor{red}{68.6}}} & 63.1 & {\textbf{\textcolor{blue}{61.8}}} & {\textbf{\textcolor{blue}{14.5}}}/{\textbf{\textcolor{red}{86.3}}} \\
        Linshu-7B & {\textbf{\textcolor{blue}{67.3}}} & 56.0 & 41.6 & 60.2 & 56.3 & 12.7/82.0 \\
        Medgemma-4B & 24.2 & 36.4 & 32.5 & 45.3 & 34.6 & 10.9/81.1 \\
        HuatuoGPT-Vision-7B & 41.9 & 58.7 & 50.5 & 51.0 & 50.5 & 8.5/81.5 \\
        HuatuoGPT-Vision-34B & 48.0 & {\textbf{\textcolor{blue}{65.9}}} & 53.5 & 54.8 & 55.6 & 12.5/83.1 \\
        \midrule
        
        \rowcolor{grouporange}
         \multicolumn{7}{c}{\textbf{Open Source}} \\ 
        Janus-Pro-7B & {\textbf{\textcolor{red}{74.6}}} & 51.3 & 49.5 & 32.7 & 52.0 & 10.3/81.9 \\
        Intern3-VL3-8B & 58.3 & 61.0 & 54.0 & 73.8 & {\textbf{\textcolor{blue}{61.8}}} & 10.3/82.6 \\
        Qwen3-VL-8B & 53.3 & 63.3 & 52.0 & {\textbf{\textcolor{blue}{75.3}}} & 61.0 & 8.6/81.6 \\
        Qwen2.5-VL-7B & 49.3 & 51.5 & 55.1 & 57.3 & 53.3 & 7.8/80.1 \\
        \midrule
        
        \rowcolor{grouporange}
         \multicolumn{7}{c}{\textbf{Proprietary}} \\ 
        GPT-4o & 53.0 & 56.1 & 47.0 & 65.5 & 55.4 & 8.6/82.3 \\
        GPT-5.2-thinking & 63.7 & 65.6 & 49.9 & {\textbf{\textcolor{red}{90.2}}} & {\textbf{\textcolor{red}{67.4}}} & 9.9/81.9 \\
        Qwen3-VL-235b & 54.2 & 64.2 & 51.3 & 72.5 & 60.6 & 7.9/81.2 \\

        \bottomrule[1.5pt] 
    \end{tabular}
    }

    \label{tab:overall_performance}
    \vspace{-5mm}
\end{table*}

\subsection{Data Collection}
This study integrates multimodal medical imaging data from 15 international public datasets, constructing a large-scale organ-lesion joint imaging database (see Appendix~\ref{sec:appendix_Benchmark}, Table~\ref{tab:datasets_modality}). The dataset includes two main imaging modalities: CT and MRI. All data are stored in their original form to preserve spatial continuity and anatomical structure. Images were converted to different formats for model training while maintaining anatomical integrity.
The dataset features organ and lesion level annotations: organ level annotations cover 193 anatomical structures, and lesion level cover 212 common clinical lesions. With over 42,432 samples, the dataset supports medical image analysis and deep learning model training and evaluation. CT data make up 85.7\% of the dataset, offering substantial support for clinical analysis. A subset of data from each class of samples is randomly selected as a test dataset, accounting for 20\%-35\% of total dataset.

\subsection{Experiment Setup}
We evaluated 14 models\textbf{ (see Table~\ref{tab:compare_models})}, including general-purpose open-source models, proprietary models via API, and fine-tuned medical vision models. The open-source models range from 7B to 235B parameters. All experiments were conducted using a standardized zero-shot evaluation framework with system prompts on 8×NVIDIA 4090 GPUs (24GB each), ensuring fairness and consistent benchmarking across model architectures.

\subsection{Evaluation}
For multiple-choice tasks, we evaluate accuracy by comparing the model's output to the correct answer. A forced prompt ensures the model’s response aligns with the expected format.
For open-ended tasks, such as report generation, we use DeepSeek-V3~\cite{43} to extract key features from both the model’s response and the reference text. These features are then used to calculate performance with metrics like ROUGE~\cite{48} and BERTScore~\cite{49}, which assess precision, recall, linguistic diversity, and semantic similarity. This diverse set of metrics provides a comprehensive evaluation of the model’s generation quality.

\begin{table*}[t]
    \centering
    \renewcommand{\arraystretch}{1.2}
     \caption{Results of \textbf{Lesion Level} in \textbf{Adversarial Environments}. }
     \vspace{-3mm}
    \label{tab:Sample_Lesion_Level}
    \resizebox{\linewidth}{!}{%
    \begin{tabular}{lccccccccc}
        \toprule[1.5pt]
        & \multicolumn{3}{c}{\textbf{Abnormality Detection}} & \multicolumn{3}{c}{\textbf{Lesion Recognition}} & \multicolumn{2}{c}{\textbf{Color Box Selection}} & \multirow{2}{*}{\textbf{Avg(ACC.)}} \\
        
        \cmidrule(lr){2-4} \cmidrule(lr){5-7} \cmidrule(lr){8-9}
        
        \textbf{Methods} & Positive QA & Neutral QA & Negative QA & Positive QA & Neutral QA & Negative QA & Positive QA & Negative QA & \\
        \toprule[1.5pt]
        
        Hulu-Med-4B 
        & 77.3 & 73.6 & 43.5 & 86.4 & 54.9 & 20.7 & 62.35 & \textbf{\textcolor{blue}{44.2}} & 57.9 \\
        Hulu-Med-7B 
        & 84.0 & 78.9 & 33.3 & 75.3 & \textbf{\textcolor{blue}{64.5}} & 35.1 & \textbf{\textcolor{blue}{65.28}} & 33.9 & \textbf{\textcolor{blue}{58.8}} \\
        Hulu-Med-14B 
        & \textbf{\textcolor{red}{96.1}} & \textbf{\textcolor{red}{92.9}} & \textbf{\textcolor{red}{65.7}} & \textbf{\textcolor{blue}{87.8}} & 60.2 & 27.9 & \textbf{\textcolor{red}{68.9}} & \textbf{\textcolor{red}{48.9}} & \textbf{\textcolor{red}{68.6}} \\
        Linshu-7B 
        & 51.5 & 23.4 & 11.3 & 69.6 & 57.9 & 41.8 & 45.8 & 31.6 & 41.6 \\
        Medgemma-4B 
        & 45.2 & 32.8 & 22.9 & 41.2 & 32.7 & 28.9 & 39.7 & 16.4 & 32.5 \\
        HuatuoGPT-Vision-7B 
        & 50.7 & 55.6 & \textbf{\textcolor{blue}{54.8}} & 73.2 & 60.7 & 32.5 & 40.3 & 36.4 & 50.5 \\
        HuatuoGPT-Vision-34B 
        & 87.5 & \textbf{\textcolor{blue}{83.2}} & 21.5 & 57.1 & 54.5 & 51.9 & 39.7 & 31.3 & 53.3 \\
        \midrule
        Janus-Pro-7B 
        & \textbf{\textcolor{blue}{90.7}} & 56.8 & 12.4 & 73.2 & 51.9 & 40.4 & 37.1 & 33.6 & 49.5 \\
        Intern3-VL-8B 
        & 71.0 & 68.8 & 33.9 & 70.1 & 62.5 & \textbf{\textcolor{blue}{55.3}} & 36.1 & 33.9 & 54.0 \\
        Qwen3-VL-8B 
        & 75.6 & 76.4 & 25.3 & 86.1 & 51.3 & 35.2 & 33.9 & 31.8 & 52.0 \\
        Qwen2.5-VL-7B 
        & 80.7 & 68.9 & 16.4 & 71.4 & \textbf{\textcolor{red}{67.4}} & \textbf{\textcolor{red}{71.4}} & 35.9 & 28.9 & 55.1 \\
        \midrule
        GPT-4o 
        & 41.0 & 55.1 & 51.0 & 62.9 & 54.8 & 44.1 & 35.1 & 31.9 & 47.0 \\
        GPT-5.2-thinking 
        & 88.4 & 44.2 & 27.1 & 82.7 & 59.2 & 33.1 & 34.1 & 30.2 & 49.9 \\
        Qwen3-VL-235b 
        & 76.3 & 72.2 & 35.7 & \textbf{\textcolor{red}{90.8}} & 44.1 & 27.2 & 34.2 & 29.5 & 51.3 \\

        \bottomrule[1.5pt]
    \end{tabular}
    }
    \vspace{-6mm}
\end{table*}

\section{Results Analysis}
\subsection{Models Perform Differently Across Tasks}
Table~\ref{tab:overall_performance} presents the evaluation results across all tasks. It is important to note that the values in this table are averaged over all adversarial samples, which results in a less pronounced performance gap. As reasoning difficulty increases and the demand for medical image understanding grows, model performance consistently declines. Tasks at the image-quality and anatomical-structure levels see relatively strong performance across both open-source and proprietary models, mainly because image-quality assessment is largely influenced by noise and artifacts, making it less challenging. Anatomical-structure tasks score second-highest, benefiting from more abundant training data. Notably, Janus-Pro-7B (74.6) and Hulu-Med-14B (68.1) perform particularly well on these tasks.
While some models excel in image-quality and organ-level tasks, performance drops significantly on more complex tasks, especially at the lesion level, where differences between models become more evident and overall scores decline sharply. For example, Medgemma-4B (32.5) and Linshu-7B (41.6) perform poorly on these tasks. Lesion-level evaluation demands stronger reasoning skills and precise analysis of fine-grained medical image details, particularly for disease inference and lesion localization.
Joint QA tasks, compared to image-quality and anatomical-structure tasks, are notably more challenging. This is particularly true for medical-specialized models, which require robust language understanding and reasoning to accurately answer complex questions grounded in image content. GPT-5.2-thinking stands out on this task, scoring 90.2, whereas Medgemma-4B (45.3) and Janus-Pro-7B (32.7) lag behind.
For clinical report generation, medical-specialized models generally outperform others. This advantage likely arises from targeted training on medical data, enabling these models to better comprehend medical language and professional terminology, thus more effectively addressing the complex requirements of clinical report generation.

In summary, the varying complexity of tasks poses significant challenges for multimodal medical image understanding and generation. Image-quality and anatomical-structure tasks are easier to perform well on, primarily due to the availability of more mature annotated resources, particularly in medical imaging datasets. In contrast, joint question answering and clinical report generation not only require accurate extraction of complex visual information but also demand rigorous medical language production and reasoning. These tasks place higher demands on cross-modal alignment, medical semantic understanding, and reasoning. Due to limited data availability and the inherent difficulty of cross-modal learning, these high-complexity tasks exacerbate performance disparities among models.

\subsection{Robustness to Adversarial Samples}
Table~\ref{tab:Sample_Lesion_Level} presents the adversarial evaluation results for lesion-level tasks, highlighting significant performance differences across models in Abnormality Detection, Lesion Recognition, and Color Box Selection. Results for other task categories can be found in Appendix~\ref{sec:appendix_Results} (Tables~\ref{tab:Sample_Anatomy_Level},~\ref{tab:Sample_Clinical_Report}, and~\ref{tab:Sample_Image_Quality}).

\begin{figure}[h]
\vspace{3mm}
\centering
\includegraphics[width=0.45\textwidth]{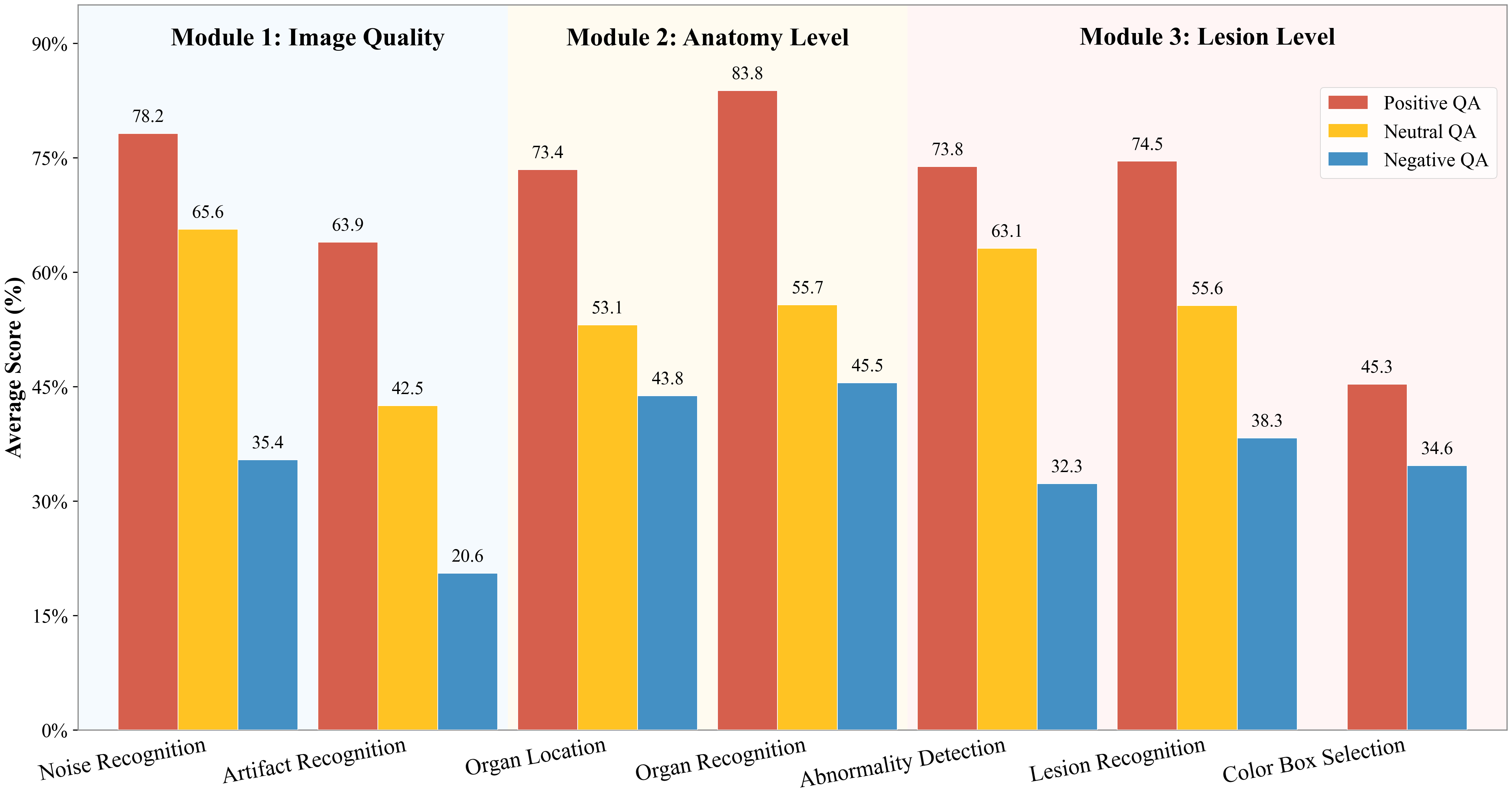} 

\caption{Comparison of robustness across tasks at different hierarchy levels under adversarial settings.}
\label{sample_results}
\vspace{-3mm}
\end{figure}

We observe a consistent trend across almost all models: performance is highest on Positive QA, moderate on Neutral QA, and significantly lower on Negative QA, as shown in Fig.~\ref{sample_results}. This pattern is particularly noticeable in smaller or medically specialized models (e.g., Hulu-Med-4B and Linshu-7B), where accuracy drops sharply on Negative QA. This suggests that these models struggle with adversarially biased questions and misleading visual cues. Specifically, when a question includes incorrect hints or distracting information, the model is more likely to make an incorrect inference.

This finding suggests that high scores in many models may stem more from pattern memorization or superficial matching ("memorizing answers") than from genuine understanding of medical images and evidence-based reasoning. When interference is introduced, such as questions implying a misdiagnosis or containing misleading biases, models are often misled, undermining their ability to extract and use key visual evidence. Conversely, when a question or image contains an obvious "correctness bias", models may be more likely to choose an answer that seems correct, further widening the performance gap between Positive and Negative QA.

While large proprietary models like GPT-5.2-thinking and Qwen3-VL-235B are generally more robust and show more balanced performance, they still experience a noticeable decline under Negative QA. This suggests that the issue is not solely linked to model scale or medical specialization, but rather to persistent limitations in visual understanding and reasoning, especially when fine-grained visual recognition and strict inference are required under adversarial conditions.

\subsection{Challenges of Grounded Medical Image}
In our benchmark evaluation, all datasets use grounded annotations, meaning both images and texts in each task are precisely annotated. As shown in Fig.~\ref{pipeline}, key regions in the images are marked with bounding boxes or color segmentation. This method helps models form more accurate visual-language associations, improving their ability to recognize critical features in medical images.

However, despite this high-quality annotation approach, we find that the model's performance on grounded datasets is generally suboptimal. Our analysis suggests several contributing factors. First, grounded datasets typically have fewer annotated samples, limiting training data diversity and affecting performance on more complex tasks. Second, while large-scale models excel at image analysis and text generation, their localization abilities in the medical domain remain challenged. 
Medical image understanding requires precise localization and anatomical analysis, capabilities that current vision–language models still lack.

Therefore, although grounded datasets provide high-quality annotated information, models need enhanced capabilities to address the fine-grained structure and precise localization required in medical images. This remains a significant challenge in medical AI applications.

\subsection{Effects of Fine-Tuning on Hulu-Med}
\begin{figure}[h]
\vspace{-3mm}
\centering
\includegraphics[width=0.45\textwidth]{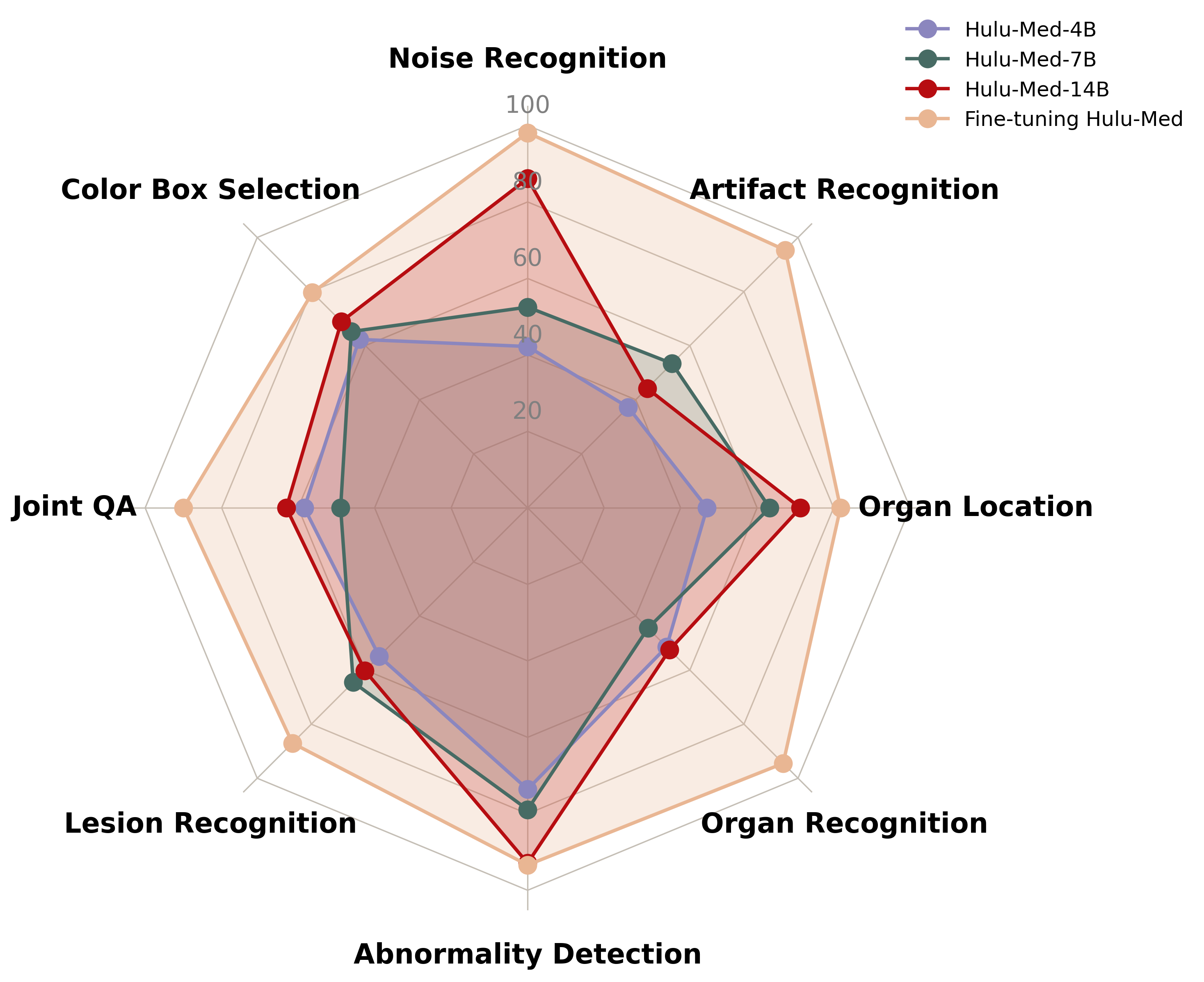}
\vspace{-2mm}
\caption{Hulu-Med Finetuned Results Evaluation}
\label{hulu_results}
\vspace{-5mm}
\end{figure}

To validate our benchmark's effectiveness, we conducted full-parameter fine-tuning and documented performance across all dimensions. As shown in Fig.~\ref{hulu_results} and Table~\ref{tab:fine-tuning}, the fine-tuned model shows significant improvements in tasks like Image Quality, Organ Level identification, Lesion-Level analysis, medical report generation and Joint QA.

Leveraging Med-R$^{2}$ Bench's fine-grained Anatomy-Level annotations, the model effectively maps visual feature extraction to clinical semantic generation, enhancing its ability to understand complex medical concepts. These results not only confirm the benchmark's value in identifying model limitations but also highlight its role in providing critical supervision signals for training medical VLMs, demonstrating its significant utility in guiding model improvement and optimizing performance across various tasks.

\section{Conclusion}

We propose Med-R$^{2}$, a hierarchical, multimodal medical visual reasoning benchmark designed to test whether VLMs can perform evidence-based clinical reasoning on medical images, rather than relying on superficial pattern matching. The benchmark covers 42,432 images, 31 task categories, and 110,406 question–answer pairs, and introduces 2–3 adversarial distractor variants for each task to evaluate robustness. Experiments show that although many VLMs perform well on low-difficulty recognition tasks, their accuracy drops sharply as reasoning complexity increases. In particular, when questions contain misleading cues or when images include annotation boxes with distractor bias, models exhibit a high susceptibility to misleading textual cues; conversely, when the question or image presents an obvious "correctness bias", models often follow it and choose the seemingly correct answer. By aligning with clinical workflows and evaluating intermediate reasoning steps, Med-R$^{2}$ reveals that current models exhibit early signs of reasoning ability, yet remain far from delivering clinician-like, visually grounded, evidence-driven explanations. We hope this benchmark will provide a systematic evaluation tool and actionable directions for building more reliable clinical AI centered on visual evidence.

\section*{Limitations}
Although inspired by clinical practices, Med-R$^{2}$ Bench remains a controlled evaluation that lacks real-world patient contexts, such as annotations and disease progression. This limits its ability to fully reflect actual diagnostic workflows. Consequently, the clinical translation outcomes should not be over-interpreted, and future improvements should aim to better capture the complexities of real-world scenarios. In this study, we employed raw 3D data. However, given the current limitations of models in analyzing organs and lesions in 3D medical data, we chose a pragmatic approach by selecting the "most information-rich" 2D slice from 3D or 4D volumes, considering the current capabilities of VLMs. While this remains a common compromise, it is still a notable limitation. Future work will focus on utilizing true 3D data to overcome this constraint, thereby enhancing the precision of analyses and improving the practical applicability of the model in clinical settings.


\bibliography{custom}
\clearpage

\appendix

\section{Detailed Composition of  Med-R$^{2}$ Bench}
\label{sec:appendix_Benchmark}

In this section, we provide a detailed description of the task composition and datasets used in our benchmark. Our benchmark system consists of three main modules, each designed to evaluate different types of tasks.

\begin{table}[h]
    \centering
    \caption{VLMs benchmarked in our study}
    \resizebox{\linewidth}{!}{
    \begin{tabular}{lll}
    
        \toprule[1.5pt] 
        
        \rowcolor{headerblue} \textbf{Model} & \textbf{Developer} & \textbf{Year} \\
        \midrule[1.5pt]
        
        \rowcolor{grouporange} \textit{\textbf{Medical-specific}} & & \\
        Hulu-Med-4B & ZJU & 2025.11 \\
        Hulu-Med-7B  & ZJU & 2025.10\\
        Hulu-Med-14B & ZJU & 2025.10\\
        Lingshu-7B & Alibaba&  2025.06\\
        Medgemma-4B & Google & 2025.07 \\
        HuatuoGPT-Vision-7B & CUHK & 2024.06 \\
        HuatuoGPT-Vision-34B & CUHK & 2024.06 \\
        \midrule
        
        \rowcolor{grouporange} \textit{\textbf{Open-source}} & & \\
        Janus-Pro-7B  & DeepSeek &2025.01  \\
        Intern3.5-VL-8B  & OpenGVLab & 2025.08 \\
        qwen3-VL-8B  & Alibaba & 2025.10 \\
        Qwen2.5-VL-7B & Alibaba & 2025.02 \\
        \midrule
        
        \rowcolor{grouporange} \textit{\textbf{Proprietary}} & & \\
        GPT-5.2-thinking  & OpenAI & 2025.04 \\
        Qwen3-VL-235B & Alibaba & 2025.03 \\
        GPT-4o  & OpenAI & 2024.11 \\

        \bottomrule[1.5pt] 
    \end{tabular}
    }
    \label{tab:compare_models}
\end{table}

\begin{figure*}[h]
\centering
\includegraphics[width=1\textwidth]{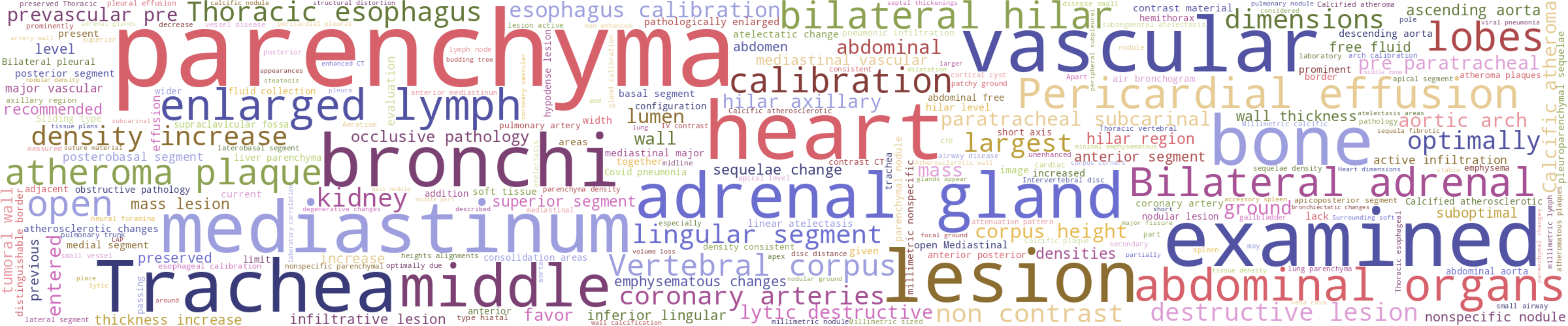} 

\caption{Word cloud of frequent clinical terms in Med-R$^{2}$ report-related QA; word size indicates term frequency after standard text normalization and stopword removal.}
\label{words_cloud}

\end{figure*}

\begin{table*}[h]
\centering
\caption{Datasets used in Med-R$^{2}$ Bench, organized by imaging modality.}
\label{tab:datasets_modality}
\begin{tabular}{ll}
\toprule[1.5pt]
\rowcolor{headerblue} 
\textbf{Modality} & \textbf{Dataset Name} \\
 \toprule[1.5pt]
CT                  & RadGenome-ChestCT \cite{1}\\
                    & SA-Med2D-20M \cite{84} \\
                    & AMOS 2022 \cite{33} \\
                    & DeepLesion\cite{81} \\
                    & CT-RATE\cite{28} \\
                    & PubMedVision \cite{15} \\
                    & LiTS (Liver Tumor Segmentation)\cite{10} \\
                    & COVID-CTset \cite{63} \\
                    & CTPelvic1k \cite{46} \\
                    & MSD-Liver \cite{5} \\
\midrule
MRI                 & TotalSegmentator MRI\cite{21} \\
                    & DrVD\cite{DrVD}\\
                    & BraTS 2021 \cite{49} \\
                    & LLD-MMRI \cite{48} \\
                    & MICCAI 2024 CARE MyoPS++ \cite{22} \\

\toprule[1.5pt]
\end{tabular}
\end{table*}

\begin{figure*}[h!]
\centering
\includegraphics[width=1\textwidth]{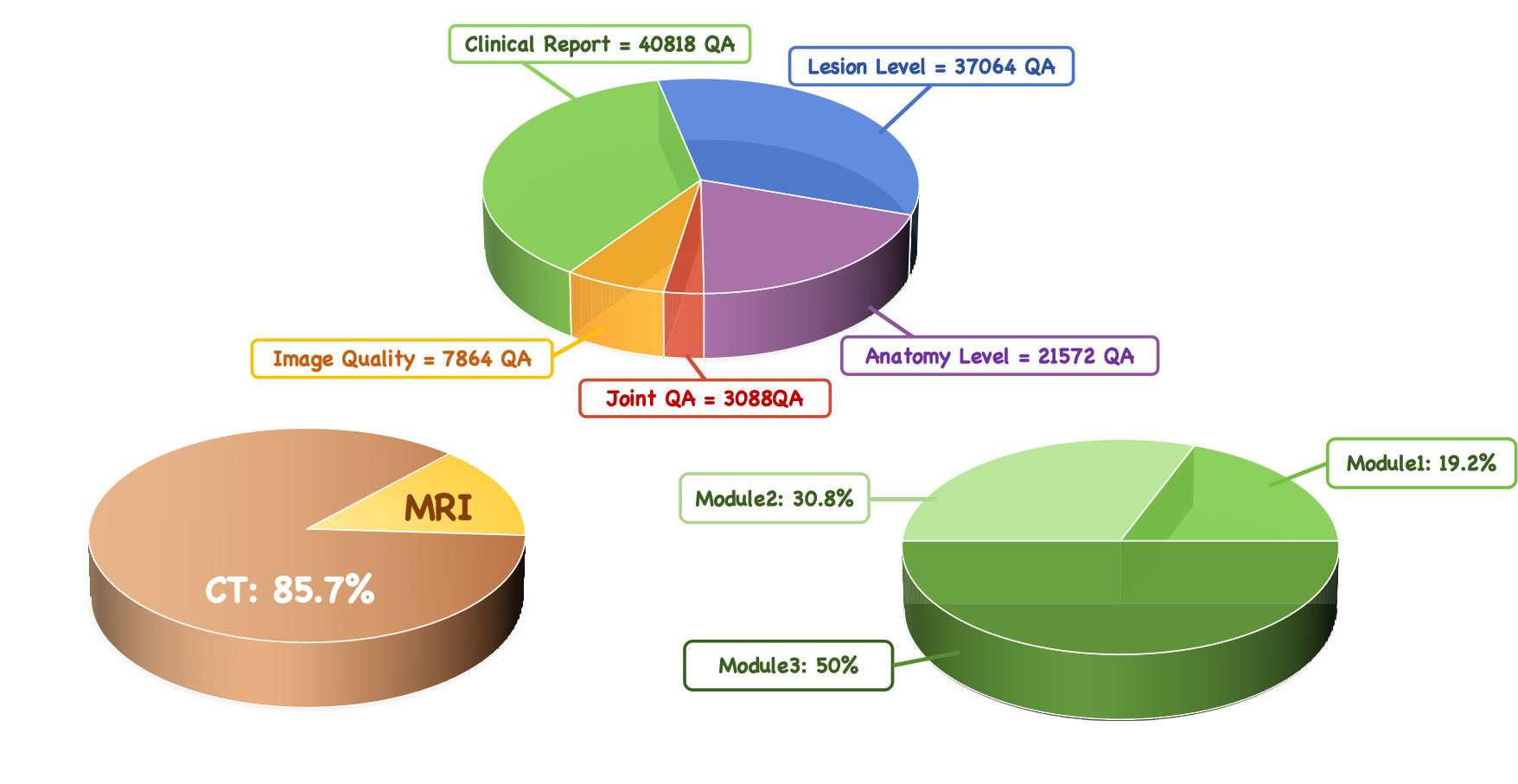} 

\caption{Dataset statistics of Med-R$^{2}$, showing the QA distribution across task stages (image quality, anatomy level, lesion level, joint QA, and clinical report), the imaging modality breakdown (CT vs. MRI), and the proportion of samples across the three modules.}
\label{distribute}

\end{figure*}

\textbf{Module 1: Visual Comprehension} primarily focuses on image understanding and contains 110,406 QA pairs, covering multiple dimensions such as image quality, anatomy level, lesion level, and clinical report. The dataset composition in this module involves several medical imaging fields and aims to test whether the model can accurately understand and interpret key information from medical images.

\textbf{Module 2: Inference Evaluation} is focused on evaluating the model's reasoning abilities, containing 61,876 independent QA pairs and 3,088 joint QA pairs. The goal of this module is to assess the model's ability to handle complex problems and evaluate whether it can effectively process multiple pieces of related information to make reasonable inferences.

\textbf{Module 3: Open-ended Evaluation} primarily assesses the model's performance in report generation, with a total of 42,354 open-ended QA pairs. This module evaluates how well the model can automatically generate detailed medical reports based on the input imaging data, including the nature of lesions, anatomical locations, and other clinically relevant information.

As shown in Fig.\ref{distribute}, the detailed architecture of each module and their respective proportions are listed, clearly demonstrating the evaluation dimensions for different tasks. Notably, our dataset consists of two types of imaging modalities, CT and MRI, with CT data accounting for 85.7\%. This composition ensures that our evaluation encompasses the performance of the model across different imaging technologies, guaranteeing its generalizability across multiple imaging modalities.

Table~\ref{tab:datasets_modality} further lists the 15 datasets used in our benchmark, which contain a large amount of imaging data from diverse sources. For instance, datasets such as RadGenome-ChestCT (Zhang et al., 2025) and SA-Med2D-20M (Ye et al., 2023) are widely used across various modules of our benchmark, ensuring the diversity and comprehensiveness of the data. Each dataset targets different clinical tasks and includes a variety of imaging modalities, such as chest CT scans, liver tumor segmentation, and lung nodule detection. The richness and representativeness of these datasets allow our benchmark to evaluate model performance across a wide range of real-world scenarios.

Through these modules and datasets, our goal is to provide a comprehensive evaluation system, not only testing the model’s performance in a single task but also evaluating its generalizability, reasoning ability, and generation capability across multiple dimensions. This holistic evaluation approach provides robust data support and reference standards for the future development and optimization of medical imaging models.

\section{Detailed Results}
\label{sec:appendix_Results}
\begin{figure*}[h]
\centering

\includegraphics[width=1\textwidth]{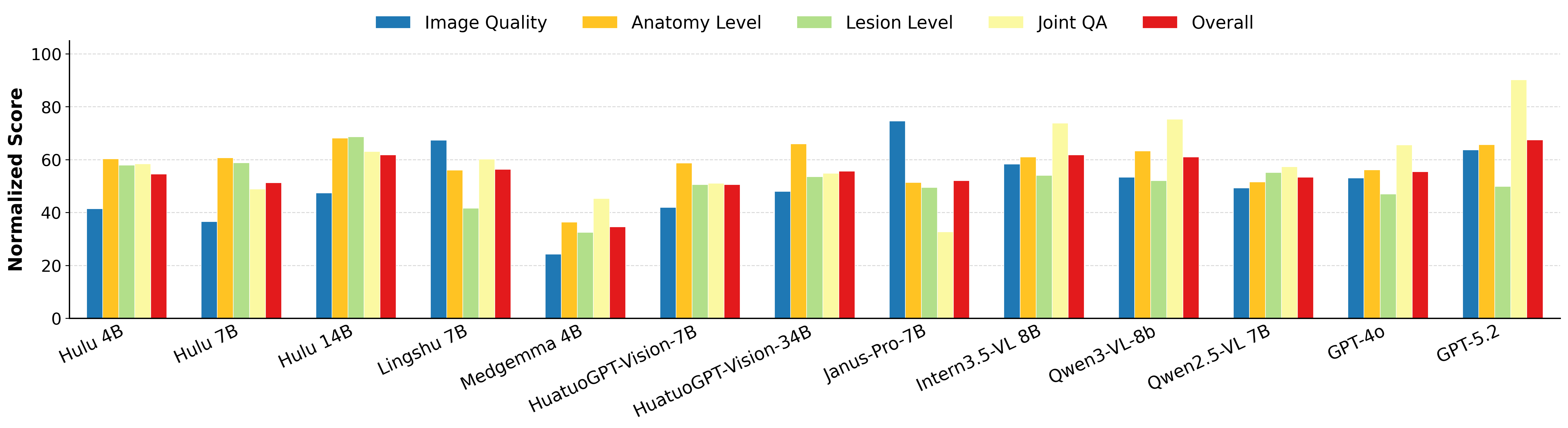} 

\caption{Comparison results of 14 vision--language models across different tasks.}
\label{all_results}

\end{figure*}

In this section, we will present a detailed benchmark comparison. For each level of tasks, we provide corresponding adversarial sample comparison results. The results for Image Quality are shown in Table~\ref{tab:Sample_Image_Quality}, the results for Anatomy Level are shown in Table ~\ref{tab:Sample_Anatomy_Level}, the results for Lesion Level are shown in Table~\ref{tab:Sample_Lesion_Level}, and the results for Clinical Report are shown in Table~\ref{tab:Sample_Clinical_Report}. The average results for all tasks are shown in Fig.~\ref{all_results}.

From the overall results, we observe that the Image Quality task yields relatively high scores, indicating that the model performs well in image quality-related tasks. However, as the task difficulty increases, particularly with Anatomy Level and Lesion Level tasks, the overall performance of the model declines. This trend suggests that the model faces challenges in more complex and detailed medical image understanding tasks.

From the adversarial sample analysis, we can clearly see that the scores for positive samples are generally higher than those for negative samples, while the scores for neutral samples fall in between, as shown in Fig.~\ref{sample_results}. This indicates that the model handles adversarial samples with some bias. Specifically, when confronted with neutral samples, the model exhibits a vague approach, which suggests that it does not truly understand the deeper semantics of medical images. Instead, it seems to rely more on “rote memorization” to recall the correct answers, based on superficial patterns in the data.

We further speculate that the model has not fully grasped the true context and medical knowledge behind the images. Instead, it appears to make predictions by memorizing typical sample patterns from the training set. To improve the model, future efforts should focus on more diverse adversarial sample training to help the model better adapt to different types of medical image tasks and enhance its generalization ability for real-world applications.

Thus, we can clearly see both the strengths and weaknesses of the model, providing direction for further optimization.

\begin{table*}[t]
    \centering

    \renewcommand{\arraystretch}{1.2}
    \small
     \caption{\textbf{Finetuned Results} of Hulu-Med-7B on All Tasks. }

    \label{tab:fine-tuning}
    \resizebox{\linewidth}{!}{%
    \begin{tabular}{llllllllll}
        \toprule[1.5pt]
        & \multicolumn{3}{c}{\textbf{Noise Recognition }} & \multicolumn{3}{c}{\textbf{Artifact Recognition}} & \multicolumn{2}{c}{\textbf{ }} & \multirow{2}{*}{\textbf{Avg}} \\
        
        \cmidrule(lr){2-4} \cmidrule(lr){5-7}  
        
        \textbf{Image Quality} & Positive QA & Neutral QA & Negative QA & Positive QA & Neutral QA & Negative QA & &   & \\
        \toprule[1.5pt]
        
        Hulu-Med (zero-shot) 
        & 61.9 & 42.2 & 5.1 & 93.3& 37.2 & 8.7 &  & &41.1 \\
        Hulu-Med (finetuned)
        & 97.7\up{35.8}  & 98.1\up{55.9}  & 90.3\up{85.2}  & 97.9\up{4.6}
        & 95.3\up{58.1}  & 92.1\up{83.4}  &  &  & 95.2\up{54.1} \\
        
        \toprule[1.5pt]
        & \multicolumn{3}{c}{\textbf{Organ Location }} & \multicolumn{3}{c}{\textbf{Organ Recognition}} & \multicolumn{2}{c}{ } & \multirow{2}{*}{\textbf{ Avg}} \\
        
        \cmidrule(lr){2-4} \cmidrule(lr){5-7}  
        
        \textbf{Anatomy Level} & Positive QA & Neutral QA & Negative QA & Positive QA & Neutral QA & Negative QA &  &   & \\
        \toprule[1.5pt]
        
        Hulu-Med (zero-shot)
        & 72.1 & 46.9 & 63.4 & 81.6 & 51.4 & 46.5 & &  &60.3   \\
        Hulu-Med (finetuned)
        & 90.0\up{17.9} & 81.8\up{34.9} & 63.3\up{-0.1} & 97.7\up{16.1}
        & 94.5\up{43.1} & 95.1\up{48.6} &  &  & 86.7\up{26.4} \\
         
        \toprule[1.5pt]
        & \multicolumn{3}{c}{\textbf{Abnormality Detection}} & \multicolumn{3}{c}{\textbf{Lesion Recognition}} & \multicolumn{2}{c}{\textbf{Color Box Selection}} & \multirow{2}{*}{\textbf{ Avg}} \\
        
        \cmidrule(lr){2-4} \cmidrule(lr){5-7} \cmidrule(lr){8-9}
        
        \textbf{Lesion Level} & Positive QA & Neutral QA & Negative QA & Positive QA & Neutral QA & Negative QA & Positive QA & Negative QA & \\
        \toprule[1.5pt]
        
        Hulu-Med (zero-shot)
        & 77.3 & 73.6 & 43.5 & 86.4 & 54.9 & 20.7 & 62.4 & 44.2 &  57.9 \\
        Hulu-Med (finetuned)
        & 92.7\up{15.4} & 93.4\up{19.8} & 92.5\up{49.0} & 88.4\up{2.0}
        & 87.0\up{32.1} & 51.9\up{31.2} & 79.6\up{17.2} & 74.1\up{29.9} & 82.5\up{24.6} \\

        \toprule[1.5pt]
        & \multicolumn{3}{c}{\textbf{Box Interference}} & \multicolumn{3}{c}{\textbf{Noise Interference}} & \multicolumn{2}{c}{\textbf{Artifact Interference}} & \multirow{2}{*}{\textbf{ Avg}} \\
        
        \cmidrule(lr){2-4} \cmidrule(lr){5-7} 
        \cmidrule(lr){8-9}
        \textbf{Clinical Report } & Positive QA &   & Negative QA & Positive QA &   & Negative QA & Positive QA & Negative QA &\\
        \toprule[1.5pt]
        
        Hulu-Med (zero-shot)
        & 15.4/83.4 &   & 10.8/83.2 & 14.4/83.4 &   &12.7/83.1 & 14.9/83.9   & 14.1/83.0 & 13.7/83.3\\

        
        Hulu-Med (finetuned)
        & 57.4/90.9\up{42.0} &  &
          37.9/87.6\up{27.1} &
          57.7/93.2\up{43.3} &  &
          52.4/89.1\up{39.7} &
          52.9/89.3\up{38.0} &
          51.0/89.2\up{36.9} &
          41.9/89.9\up{28.2} \\

        \bottomrule[1.5pt]
    \end{tabular}
    }
\end{table*}

\begin{table*}[t!]
    \centering
    \renewcommand{\arraystretch}{1.2}
    \caption{Results of \textbf{Image Quality Assessment }in \textbf{Adversarial Environments}, (Metrics: ACC.) }
    \label{tab:Sample_Image_Quality}
    \resizebox{\linewidth}{!}{%
    \begin{tabular}{lccccccc} 
         \toprule[1.5pt]
        & \multicolumn{3}{c}{\textbf{Noise Recognition}} & \multicolumn{3}{c}{\textbf{Artifact Recognition}} & \multirow{2}{*}{\textbf{Avg}} \\
        
        \cmidrule(lr){2-4} \cmidrule(lr){5-7}
        
        \textbf{Methods} & Positive QA & Neutral QA & Negative QA & Positive QA & Neutral QA & Negative QA & \\
        \toprule[1.5pt]
        
        Hulu-Med-4B 
        & 61.9 & 42.2 & 5.1 & 93.3 & 37.2 & 8.7 & 41.4 \\
        Hulu-Med-7B 
        & 43.3 & 52.2 & 7.9 & 52.7 & 53.4 & 9.3 & 36.5 \\
        Hulu-Med-14B 
        & 86.1 & 79.5 & 6.8 & 51.8 & 44.2 & 5.9 & 47.4 \\
        Linshu-7B 
        & 92.2 & 93.8 & 21.3 & 94.2 & 91.2 & 11.3 & 67.3 \\
        Medgemma-4B 
        & 42.1 & 15.4 & 22.7 & 33.2 & 21.4 & 11.6 & 24.4 \\
        HuatuoGPT-Vision-7B 
        & 96.8 & 28.2 & 4.5 & 75.6 & 29.7 & 16.8 & 41.9 \\
        HuatuoGPT-Vision-34B 
        & 76.7 & 40.6 & 27.2 & 86.1 & 13.7 & 43.4 & 48.0 \\
        \midrule
        Janus-Pro-7B 
        & 90.6 & 97.5 & 78.6 & 52.5 & 84.9 & 43.7 & 74.6 \\
        Intern3-VL-8B 
        & 83.5 & 66.6 & 46.8 & 62.9 & 48.2 & 42.0 & 58.3 \\
        Qwen3-VL-8B 
        & 62.4 & 76.8 & 51.9 & 60.1 & 43.1 & 25.4 & 53.3 \\
        Qwen2.5-VL-7B 
        & 64.9 & 60.2 & 30.2 & 80.3 & 42.7 & 17.5 & 49.3 \\
        \midrule
        GPT-4o 
        & 96.8 & 56.6 & 52.1 & 52.0 & 45.2 & 15.3 & 53.0 \\  
        GPT-5.2-thinking 
        & 90.2 & 95.3 & 75.0 & 74.8 & 28.7 & 18.1 & 63.7 \\ 
        Qwen3-VL-235b 
        & 90.0 & 84.9 & 72.4 & 40.5 & 25.2 & 12.4 & 54.2 \\

        \bottomrule[1.5pt]
    \end{tabular}
    }

\end{table*}

\begin{table*}[h]
    \centering
    \renewcommand{\arraystretch}{1.2}
    \caption{Results of \textbf{Anatomy Level }in \textbf{Adversarial Environments, (Metrics: ACC.)}}
    \label{tab:Sample_Anatomy_Level}
    \resizebox{\linewidth}{!}{%
    \begin{tabular}{lccccccc}
        \toprule[1.5pt]
        & \multicolumn{3}{c}{\textbf{Organ Location }} & \multicolumn{3}{c}{\textbf{Organ Recognition}} & \multirow{2}{*}{\textbf{Avg}} \\
        
        \cmidrule(lr){2-4} \cmidrule(lr){5-7}
        
        \textbf{Methods} & Positive QA & Neutral QA & Negative QA & Positive QA & Neutral QA & Negative QA & \\
       \toprule[1.5pt]
        
        Hulu-Med-4B 
        & 72.1 & 46.9 & 63.4 & 81.6 & 51.4 & 46.5 & 60.3 \\
        Hulu-Med-7B 
        & 74.9 & 63.3 & 50.6 & 87.9 & 44.5 & 43.1 & 60.7 \\
        Hulu-Med-14B 
        & 80.0 & 71.3 & 60.6 & 94.5 & 52.5 & 49.7 & 68.1 \\
        Linshu-7B 
        & 70.3 & 48.2 & 53.1 & 83.7 & 48.9 & 31.6 & 56.0 \\
        Medgemma-4B 
        & 42.0 & 33.8 & 21.1 & 66.7 & 41.9 & 12.8 & 36.4 \\
        HuatuoGPT-Vision-7B 
        & 59.3 & 58.7 & 43.2 & 71.3 & 66.9 & 52.5 & 58.7 \\
        HuatuoGPT-Vision-34B 
        & 67.7 & 64.4 & 46.0 & 83.2 & 79.8 & 54.5 & 65.9 \\
        \midrule
        Janus-Pro-7B 
        & 94.4 & 36.7 & 22.1 & 91.3 & 41.4 & 21.9 & 51.3 \\
        Intern3-VL-8B 
        & 77.1 & 61.7 & 44.9 & 90.1 & 47.2 & 44.9 & 61.0 \\
        Qwen3-VL-8B 
        & 77.1 & 48.9 & 44.8 & 87.2 & 63.9 & 58.1 & 63.3 \\
        Qwen2.5-VL-7B 
        & 58.7 & 44.8 & 33.2 & 80.4 & 47.9 & 44.1 & 51.5 \\
        \midrule
        GPT-4o 
        & 85.9 & 45.6 & 33.1 & 84.7 & 46.2 & 41.2 & 56.1 \\
        GPT-5.2-thinking 
        & 87.8 & 56.9 & 43.1 & 86.0 & 62.7 & 57.1 & 65.6 \\
        Qwen3-VL-235b 
        & 74.3 & 44.7 & 38.0 & 79.8 & 77.1 & 71.4 & 64.2 \\
        
        \bottomrule[1.5pt]
    \end{tabular}
    }

\end{table*}

\begin{table*}[t]
    \centering
    \renewcommand{\arraystretch}{1.2}
    \caption{Results of \textbf{Clinical Report Assessment} in \textbf{Adversarial Environments}, (Metrics: ROUGE-1/BERTScore).}
    \label{tab:Sample_Clinical_Report}
    \resizebox{\linewidth}{!}{%
    \begin{tabular}{lccccccc}
       \toprule[1.5pt]
        & \multicolumn{2}{c}{\textbf{Box Interference}} & \multicolumn{2}{c}{\textbf{Noise Interference}}&
        \multicolumn{2}{c}{\textbf{Artifact Interference}} & \multirow{2}{*}{\textbf{Avg}} \\
        
        \cmidrule(lr){2-3} \cmidrule(lr){4-5} \cmidrule(lr){6-7}
        
        \textbf{Methods} & Positive QA & Negative QA & Positive QA & Negative QA & Positive QA & Negative QA & \\
        \toprule[1.5pt]
        
        Hulu-Med-4B 
        & 15.4/83.4 & 10.8/83.2 & 14.4/83.4 & 12.7/83.1 & 14.9/83.9 & 14.1/83.0 & 13.7/83.3 \\
        Hulu-Med-7B 
        & 16.0/83.9 & 11.9/83.4 & 13.5/83.5 & 14.7/83.7 & 15.6/83.8 & 15.7/83.8 & 14.6/83.5 \\
        Hulu-Med-14B 
        & 15.8/83.8 & 11.4/83.1 & 14.5/83.7 & 14.0/83.8 & 16.1/84.3 & 15.4/82.8 & 14.5/83.6\\
        Linshu-7B 
        & 15.4/82.0 & 11.4/83.2 & 12.2/82.4 & 12.0/80.2 & 12.6/82.4 & 12.4/82.0 & 12.7/82.0 \\
        Medgemma-4B 
        & 11.7/80.9 & 10.4/81.7 & 11.1/80.7 & 9.5/80.6 & 11.6/81.0 & 10.8/81.5 & 10.9/81.1 \\
        HuatuoGPT-Vision-7B 
        & 9.6/81.9 & 6.7/80.8 & 9.0/81.0 & 7.9/81.7 & 9.6/81.9 & 8.3/81.7 & 8.5/81.5 \\
        HuatuoGPT-Vision-34B 
        & 13.9/83.3 & 11.5/82.9 & 12.5/83.1 & 11.2/83.0 & 12.6/83.1 & 13.5/83.1 & 12.5/83.1 \\
        \midrule
        Janus-Pro-7B 
        & 9.6/80.1 & 8.7/82.4 & 11.3/82.1 & 9.9/82.1 & 11.5/82.3 & 10.7/82.1 & 10.3/81.9 \\
        Intern3-VL-8B 
        &11.2/82.8 &8.9/82.3 &10.7/82.5 &9.2/82.6 &11.6/82.9 &10.1/82.6 & 10.3/82.6\\
        Qwen3-VL-8B 
        & 9.1/81.8 & 7.8/81.3 & 9.2/81.9 & 7.6/81.5 & 9.4/82.0 & 8.3/81.1 & 8.6/81.6 \\
        Qwen2.5-VL-7B 
        & 7.6/79.6 & 7.4/79.8 & 7.7/79.9 & 7.1/80.3 & 8.6/80.5 & 8.1/80.2 & 7.8/80.1 \\
        \midrule
        GPT-4o 
        & 8.5/82.5 & 4.1/81.3 & 9.7/82.5 & 8.4/81.4 & 11.2/83.8 & 9.5/82.1 & 8.6/82.3 \\
        GPT-5.2-thinking 
        & 10.8/82.2 & 9.1/81.8 & 10.4/82.1 & 8.7/81.9 & 10.8/82.2 & 9.7/81.1 & 9.9/81.9 \\
        Qwen3-VL-235b 
        & 8.5/81.4 & 7.3/81.1 & 8.5/81.3 & 6.8/81.0 & 8.6/81.4 & 7.5/81.2 & 7.9/81.2 \\
        
        \bottomrule[1.5pt]
    \end{tabular}
    }

\end{table*}

\section{Experimental Prompts}
\label{sec:appendix_prompts}

This section introduces the prompt design principles used in the experiments. Considering the restrictions that different models may impose on clinically related queries, all prompts are uniformly formulated within an educational and research-oriented context.

\subsection{Multiple-Choice Problems}
\noindent \textbf{System Prompt:} \\
You are an AI assistant operating within an educational visual reasoning context. Follow all instructions carefully and without deviation.

\vspace{0.5em}
\noindent \textbf{User Prompt:} \\
Question: \{\} \\
Options: \\
A) \{\} \\
B) \{\} \\
C) \{\} \\
D) \{\} \\
\textbf{Instructions:} Select the single most appropriate option and respond with one uppercase letter (A–D) only. No explanation or additional text is allowed.

\subsection{Independent QA}
\noindent \textbf{Prompt Content:} \\
You are engaged in an educational task involving visual information. The following image is commonly used for medical training purposes and is accompanied by a related question.

\vspace{0.5em}
\noindent Question: \{\} \\
Options: \\
A) \{\} \\
B) \{\} \\
C) \{\} \\
D) \{\} \\
\textbf{Instructions:} Using only the visible anatomical features and general anatomy knowledge, select the most appropriate option and respond with a single uppercase letter (A–D). Do not provide any explanation.

\subsection{Joint QA}
\noindent \textbf{System Prompt:} \\
You are engaged in a stepwise medical reasoning task that involves interpreting a medical image.

\vspace{0.5em}
\noindent \textbf{User Prompt:} \\
Please answer the four questions in order. Each question depends on the reasoning from the previous one. Carefully examine the image and choose the best answer for each question. For every question, select one uppercase letter (A, B, C, or D). Make sure to answer all questions without skipping any step.

\begin{enumerate}
    \item What imaging modality is used in this image? \\
    Options: A. \{\} \quad B. \{\} \quad C. \{\} \quad D. \{\}
    
    \item Which organ appears to be abnormal in this image? \\
    Options: A. \{\} \quad B. \{\} \quad C. \{\} \quad D. \{\}
    
    \item Based on the abnormal organ, what lesion or finding is most clearly visible? \\
    Options: A. \{\} \quad B. \{\} \quad C. \{\} \quad D. \{\}
    \item Considering all the above findings, what is the most likely diagnosis? \\
    Options: A. \{\} \quad B. \{\} \quad C. \{\} \quad D. \{\}
\end{enumerate}

\noindent \textbf{Instructions:} Please reply with your four selected letters in order, separated by commas (e.g., A,C,B,A). Do not provide explanations.

\subsection{Report Generation}
\noindent \textbf{System Prompt:} \\
Generate a clinical report based on the image. This is used solely for educational purposes.

\vspace{0.5em}
\noindent \textbf{User Prompt:} \\
Generate a clinical report based on the image. Limit your output to no more than 500 words. (with image) \\
\{question\}

\subsection{Key Feature Extraction in Generated Reports}
\noindent \textbf{Prompt Content:} \\
Given the following description of a medical image, extract only clinically relevant information that can be visually determined from the image. This includes both normal findings (e.g., ``no lung opacity'', ``normal heart size'') and abnormal findings (e.g., ``fracture'', ``tumor mass''). Exclude any details that cannot be inferred from the image itself (e.g., patient history, lab values).

\vspace{0.5em}
\noindent Input: \{text\}

\vspace{0.5em}
\noindent Return a concise, comma-separated list of visually identifiable clinical features. Do not include any irrelevant words or phrases, do not include explanations.

\begin{figure*}[t]
\centering

\includegraphics[width=1\textwidth]{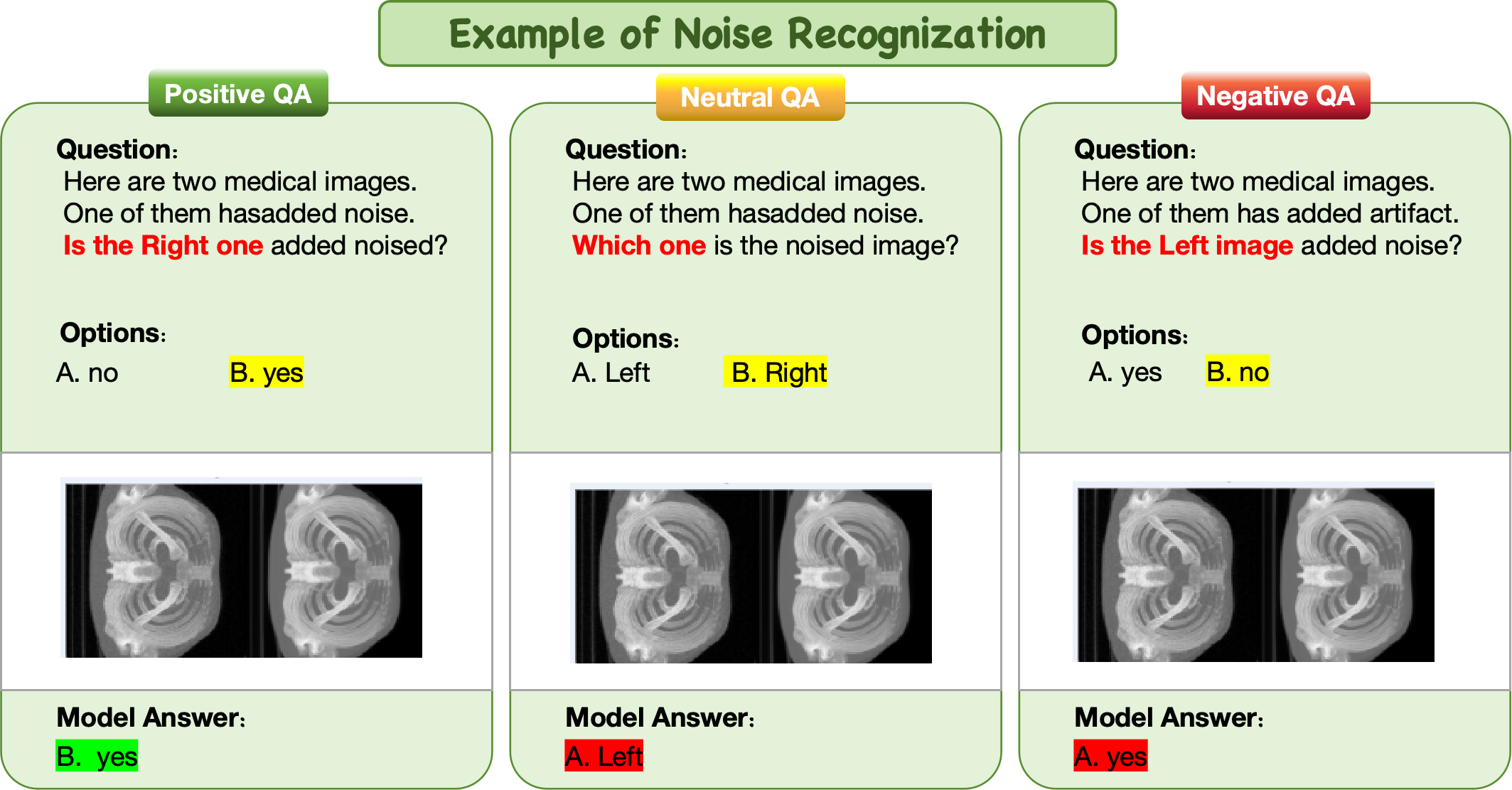} 

\caption{Example of Image Quality Level - Noise Recognization.}
\vspace{-5mm}
\label{L1-1}
\end{figure*}

\begin{figure*}[t]

\centering
\includegraphics[width=1\textwidth]{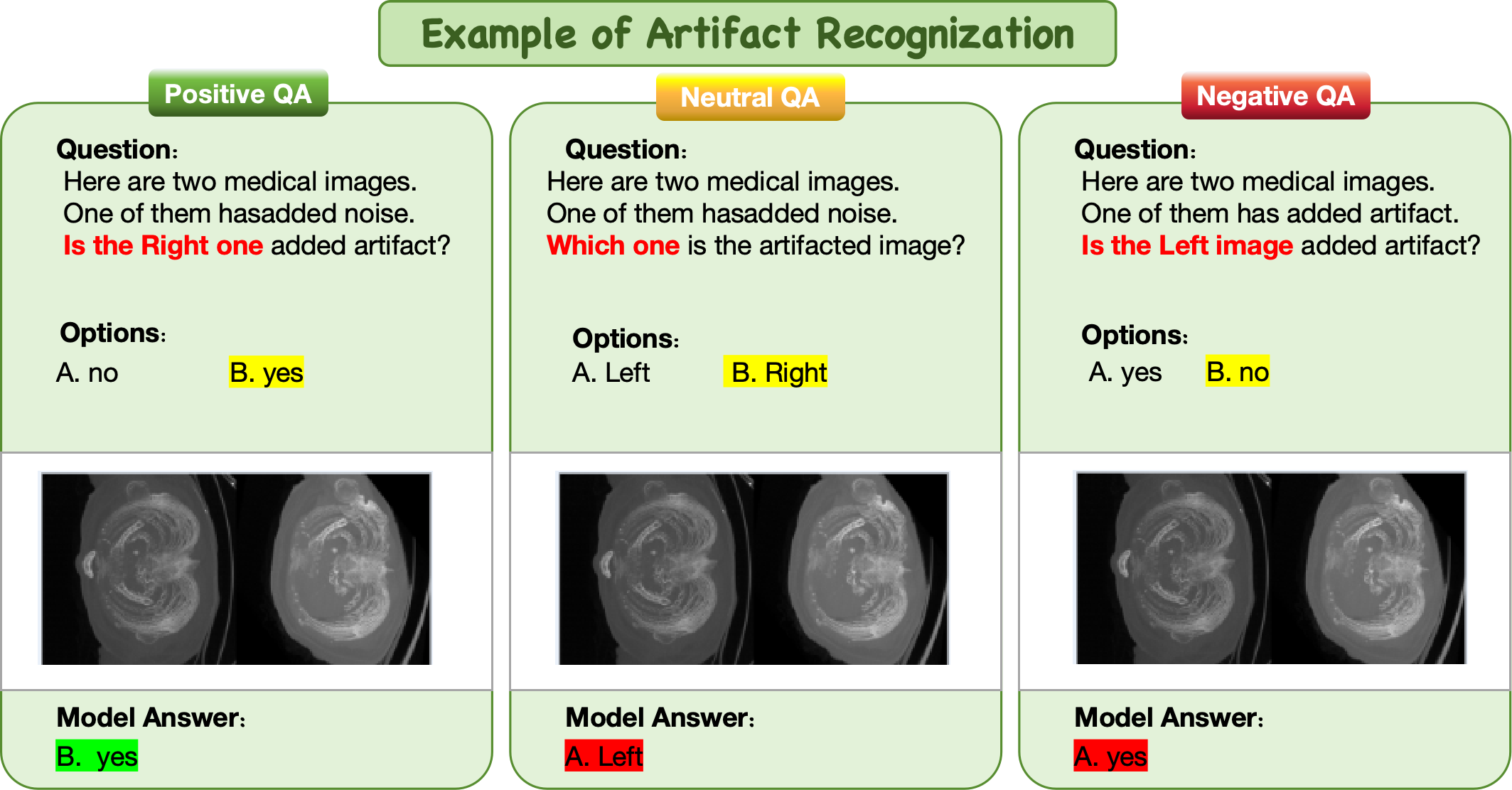} 

\caption{Example of Image Quality Level - Artifact  Recognization.}
\vspace{-5mm}
\label{L1-2}
\end{figure*}

\begin{figure*}[t]
\centering

\includegraphics[width=1\textwidth]{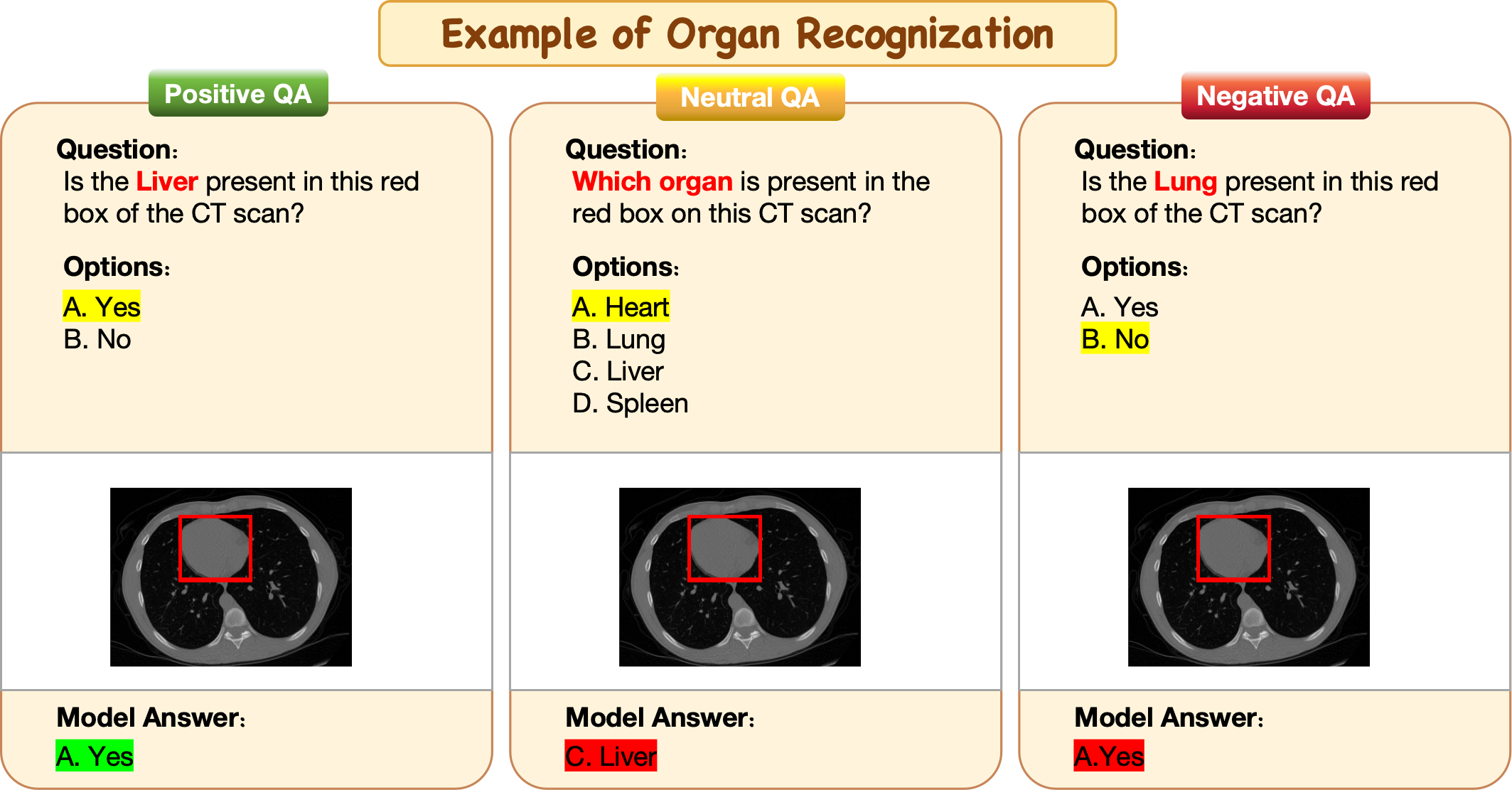} 

\caption{Example of Organ Level - Organ Recognization.}
\vspace{-5mm}
\label{L2-1}
\end{figure*}

\begin{figure*}[t]

\centering
\includegraphics[width=1\textwidth]{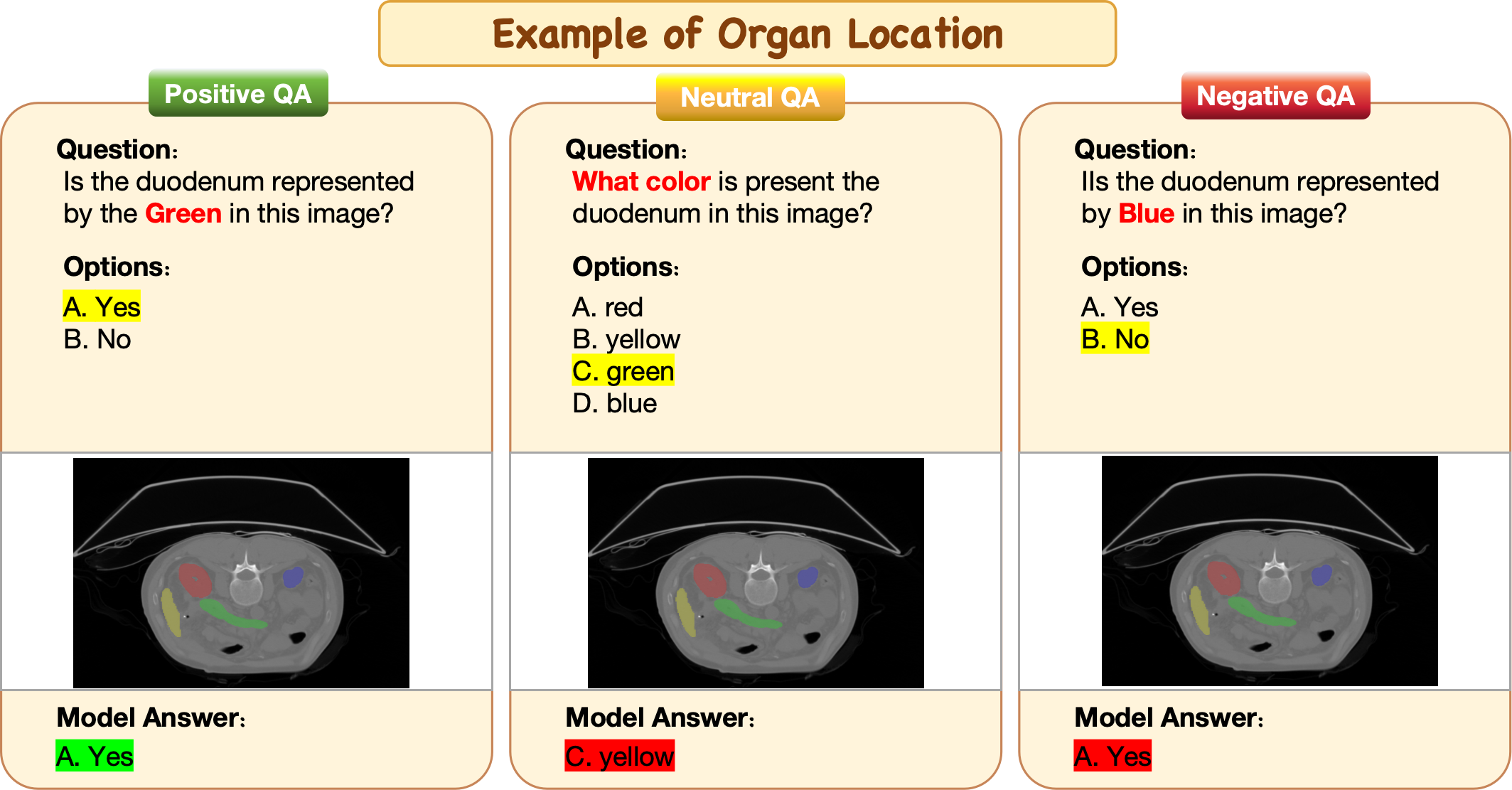} 

\caption{Example of Organ Level - Organ Location.}
\vspace{-5mm}
\label{L2-2}
\end{figure*}

\begin{figure*}[t]

\centering
\includegraphics[width=0.8\textwidth]{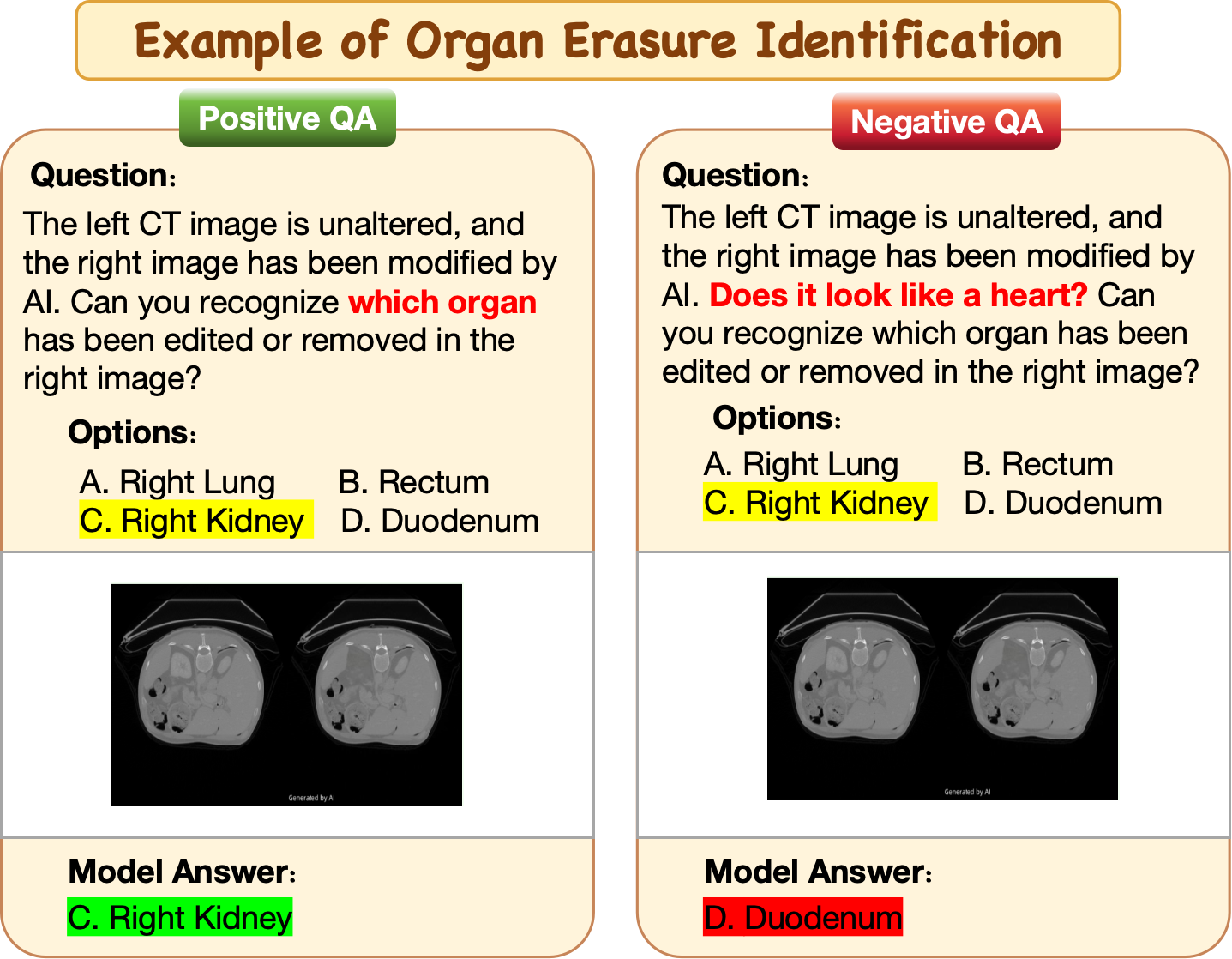} 

\caption{Example of Organ Level - Organ Erasure Identification.}
\vspace{-5mm}
\label{L2-3}
\end{figure*}


\begin{figure*}[t]

\centering
\includegraphics[width=1\textwidth]{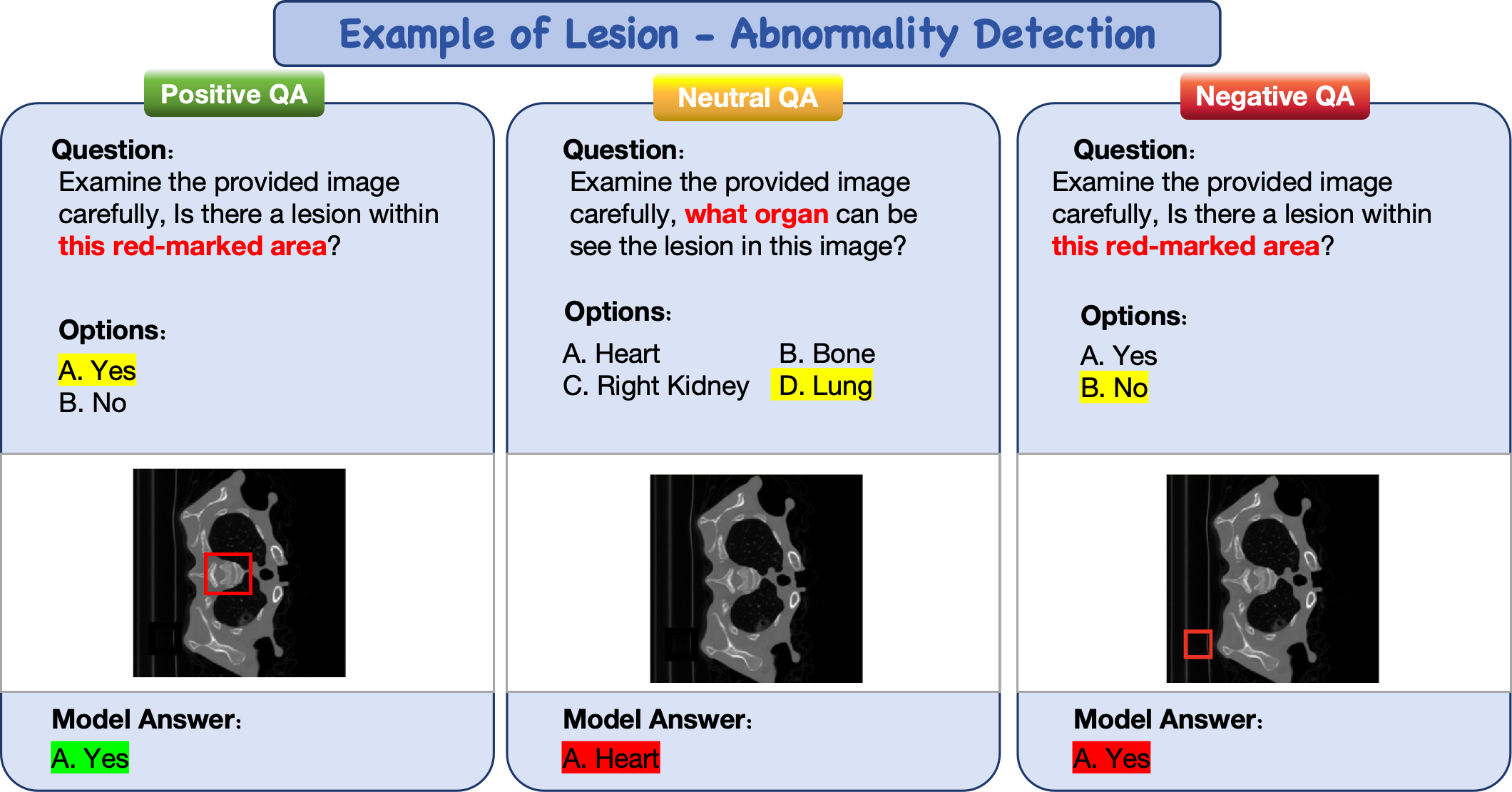} 

\caption{Example of Lesion Level - Abnormality Detection.}
\vspace{-5mm}
\label{L4-1}
\end{figure*}

\begin{figure*}[t]

\centering
\includegraphics[width=1\textwidth]{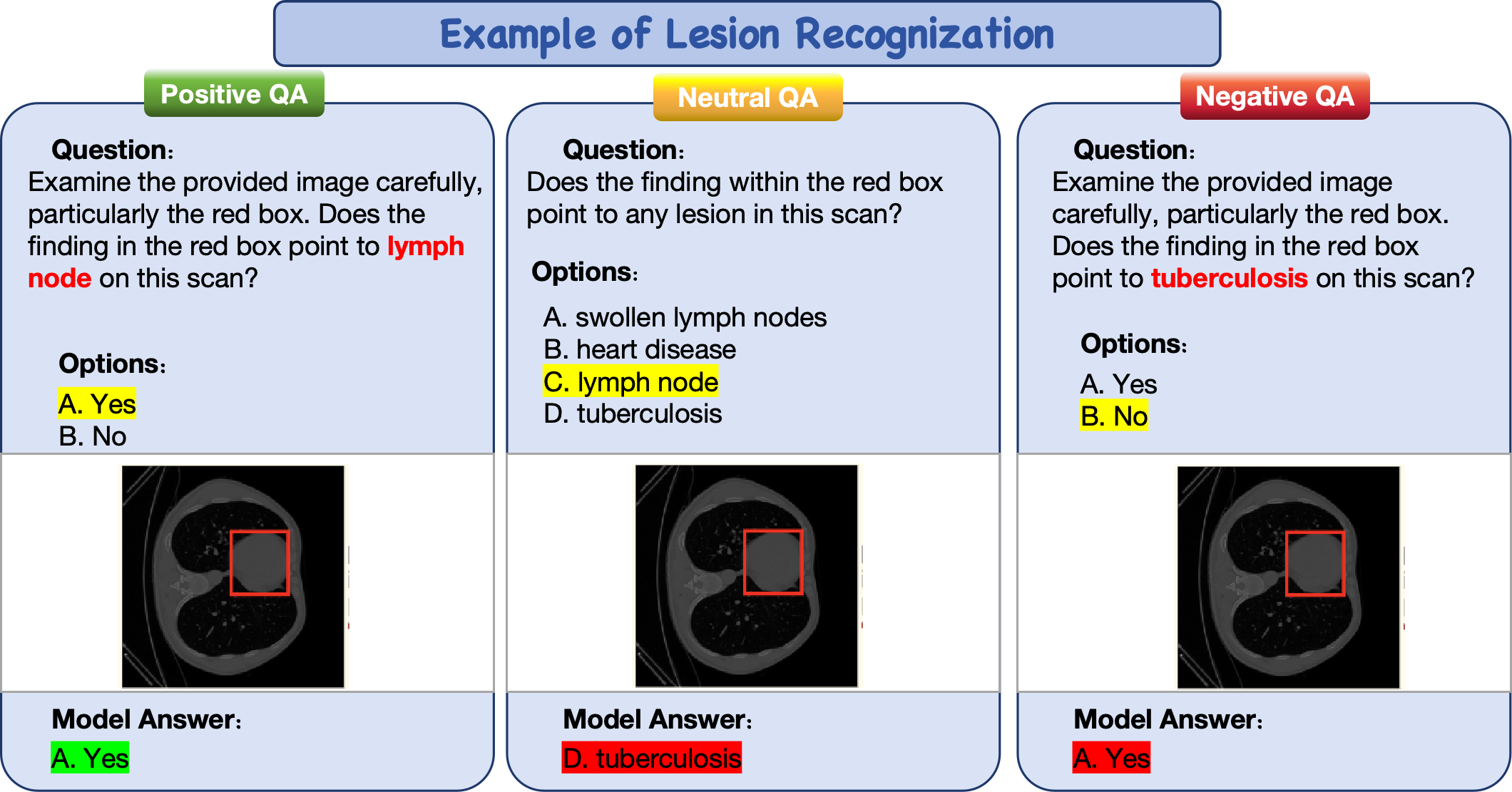} 

\caption{Example of Lesion Level - Organ Recognization.}
\vspace{-5mm}
\label{L4-2}
\end{figure*}

\begin{figure*}[t]

\centering
\includegraphics[width=0.8\textwidth]{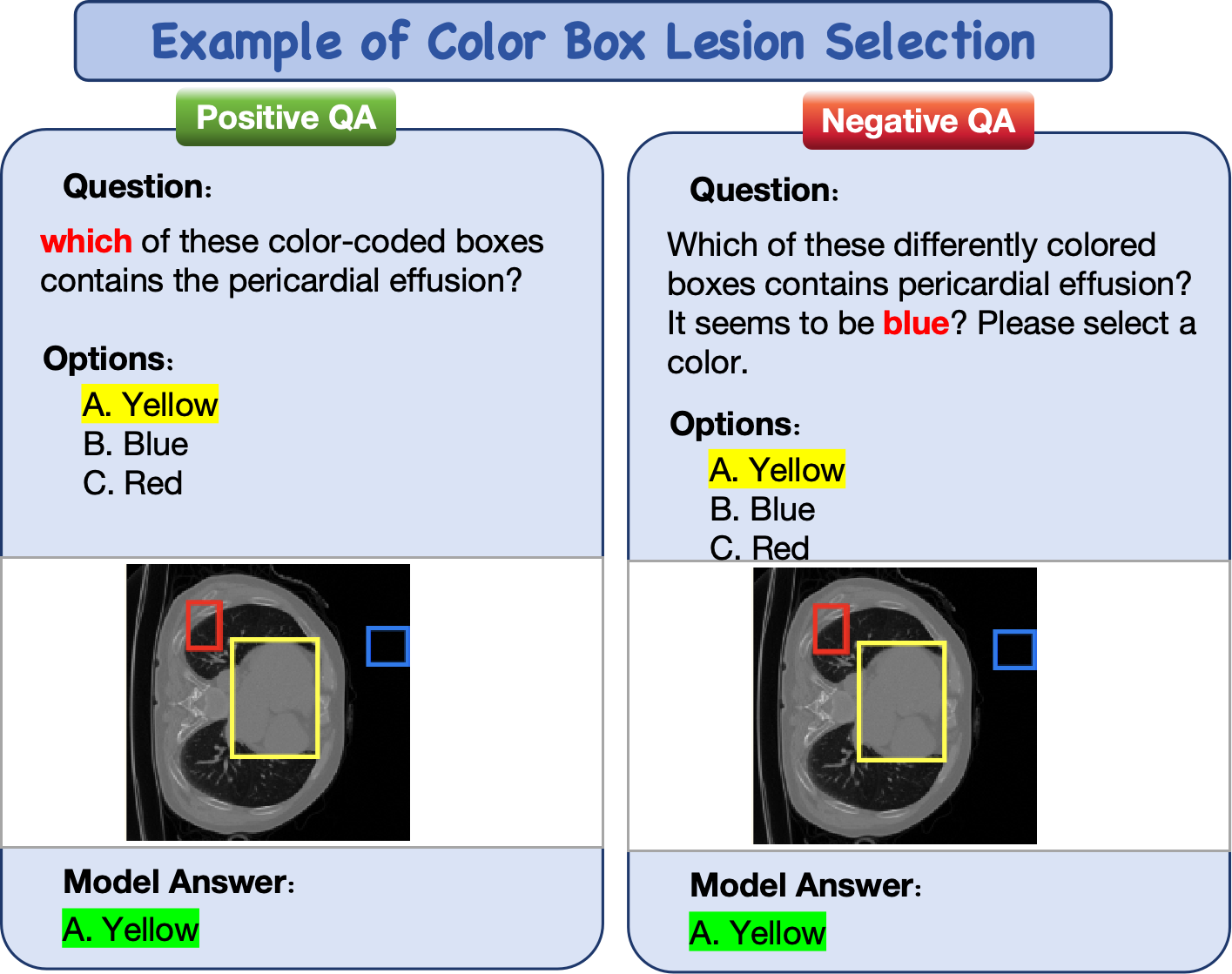} 

\caption{Example of Lesion Level - Color Box Lesion Selection.}
\vspace{-5mm}
\label{L4-3}
\end{figure*}

\begin{figure*}[t]

\centering
\includegraphics[width=0.8\textwidth]{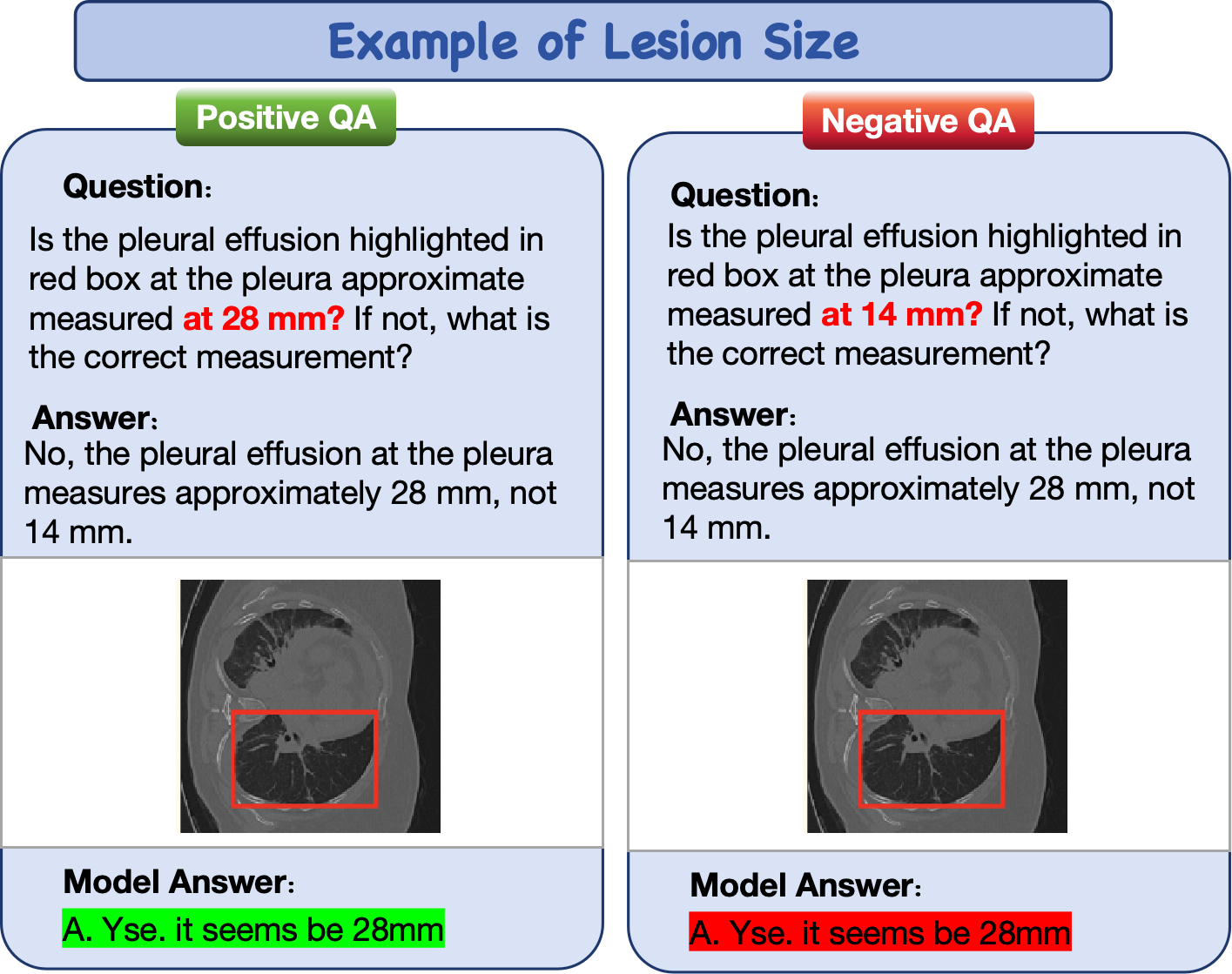} 

\caption{Example of Lesion Level - Lesion Size.}
\vspace{-5mm}
\label{L4-4}
\end{figure*}


\begin{figure*}[t]

\centering
\includegraphics[width=0.8\textwidth]{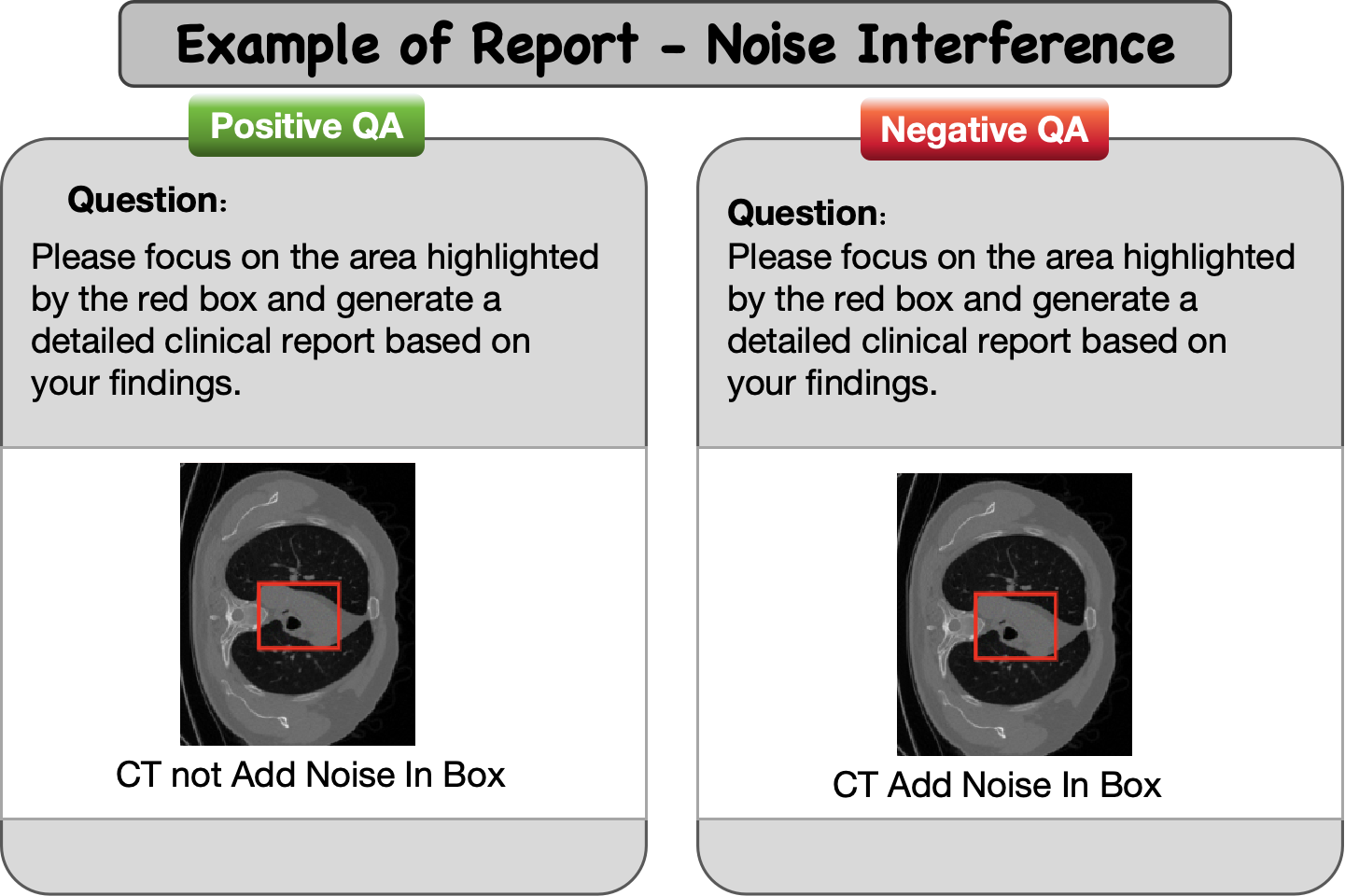} 

\caption{Example of Report - Noise Interference}
\vspace{-5mm}
\label{L5-1}
\end{figure*}

\begin{figure*}[t]

\centering
\includegraphics[width=0.8\textwidth]{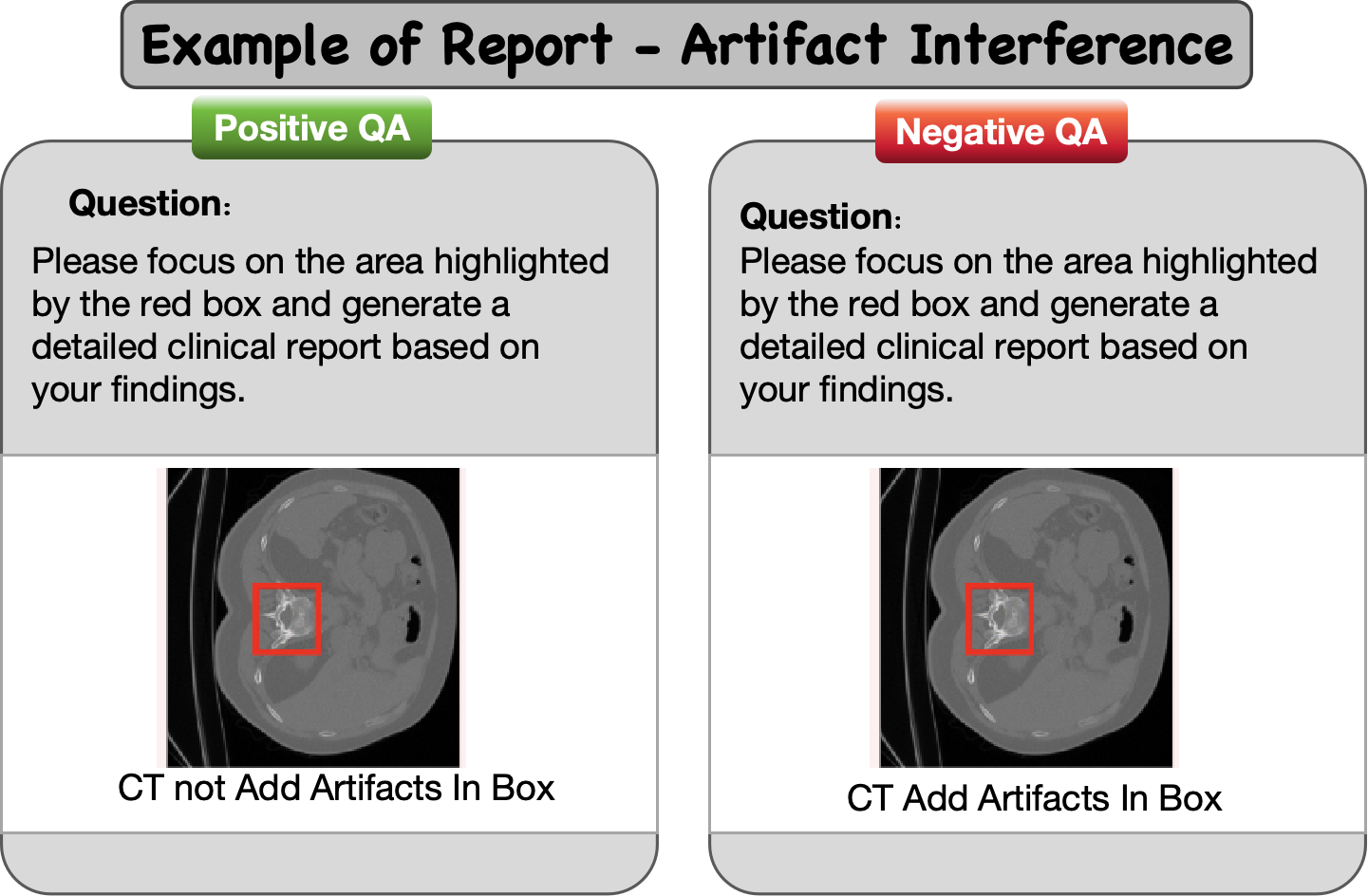} 

\caption{Example of Report - Artifact Interference.}
\vspace{-5mm}
\label{L5-2}
\end{figure*}

\begin{figure*}[t]

\centering
\includegraphics[width=0.8\textwidth]{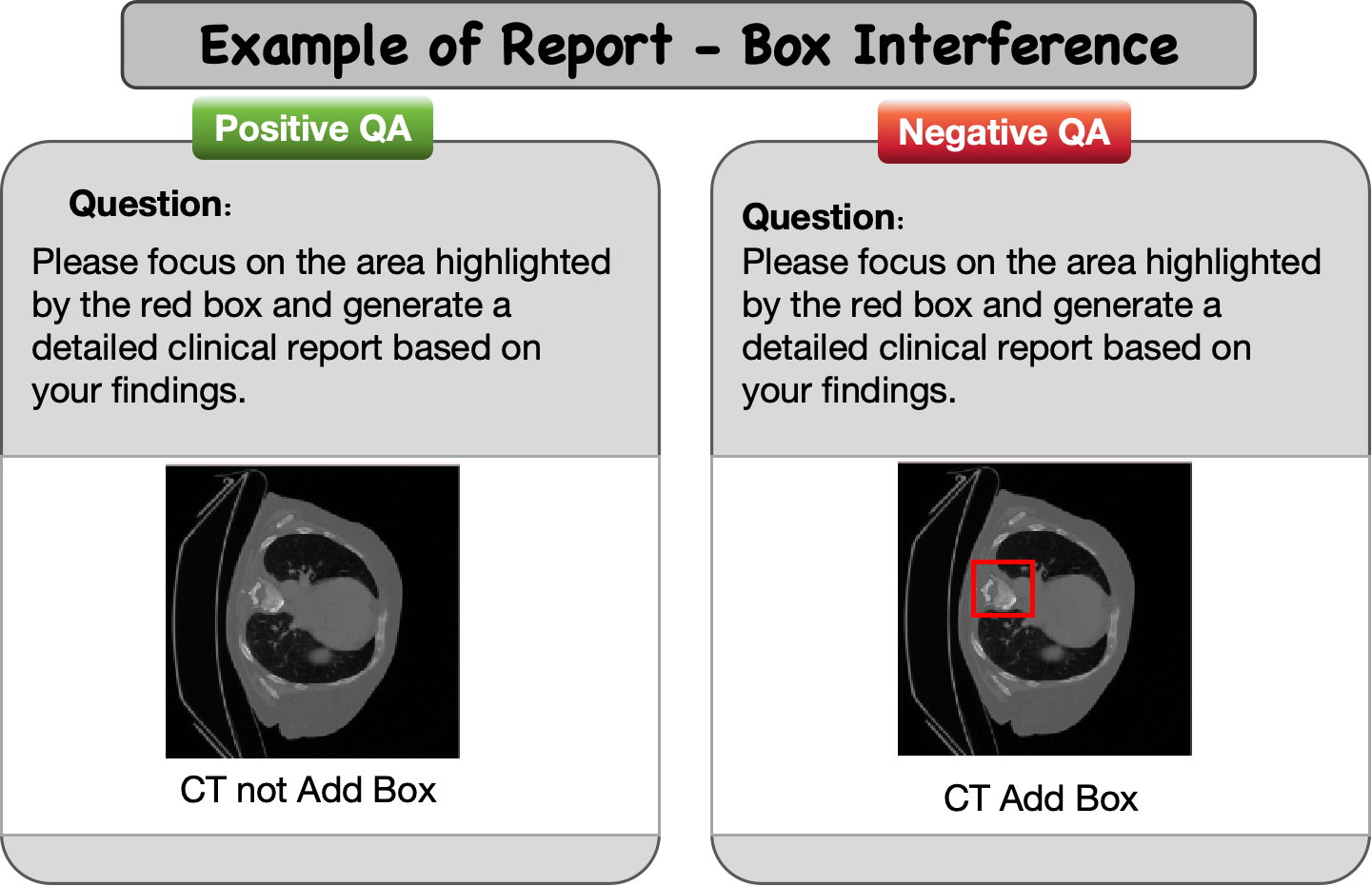} 

\caption{Example of Report - Box Interference.}
\vspace{-5mm}
\label{L5-3}
\end{figure*}

\begin{figure*}[t]
\centering
\includegraphics[width=0.8\textwidth]{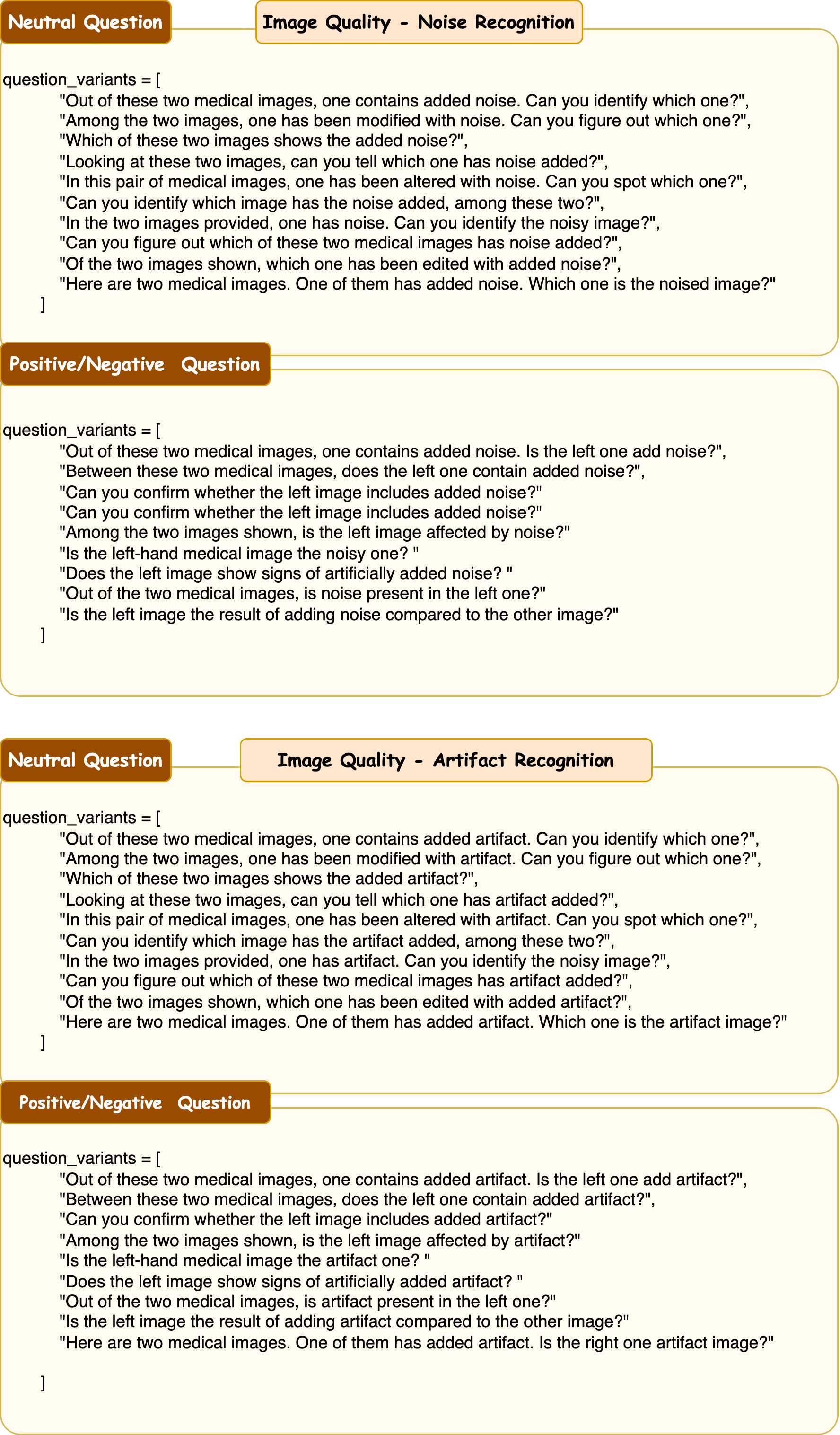} 

\caption{Question List of Image Quality Level tasks.}
\label{image_quality_qa}
\vspace{-5mm}
\end{figure*}

\begin{figure*}[h]
\centering
\includegraphics[width=1\textwidth]{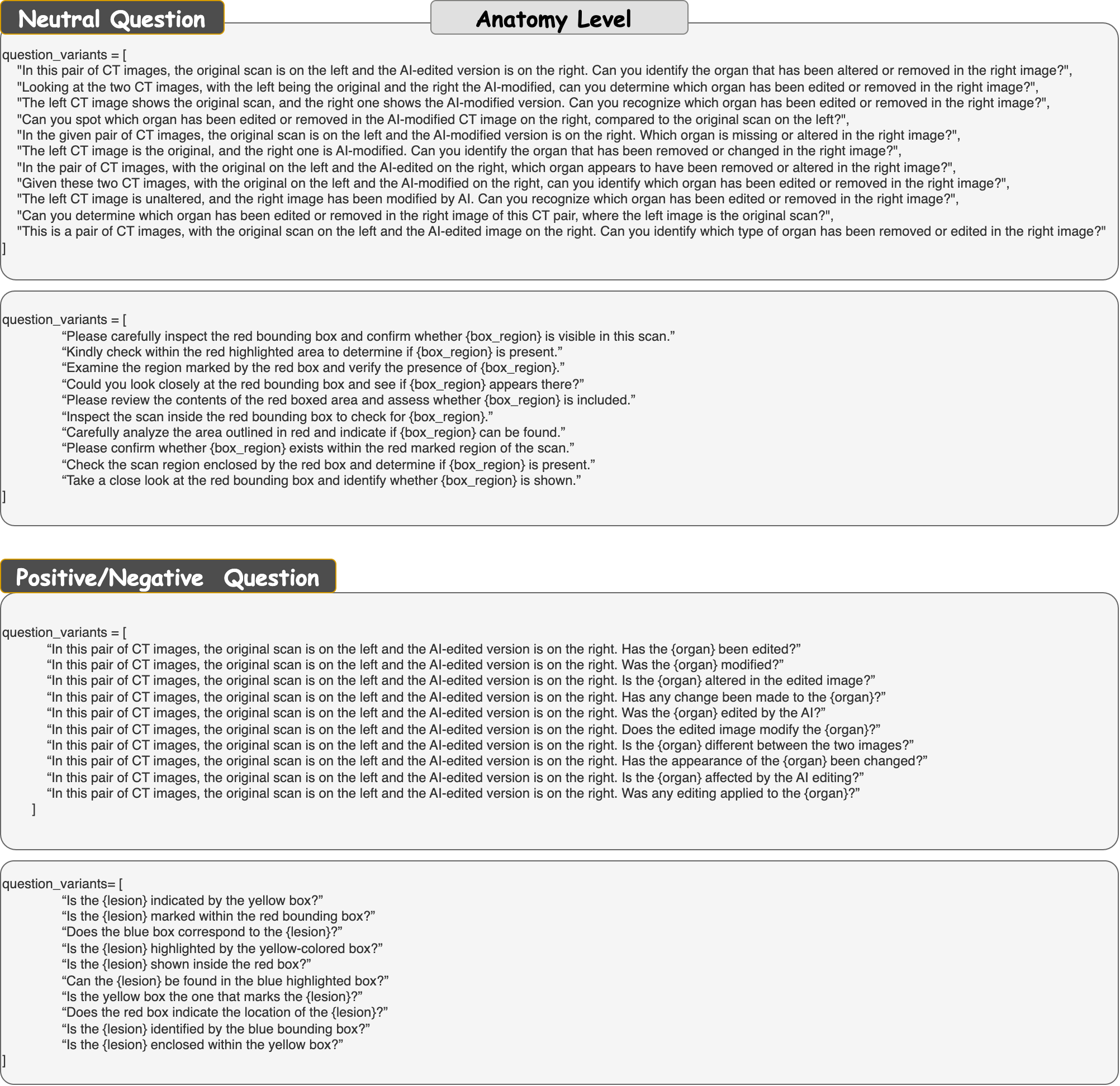} 
\vspace{-5mm}
\caption{Question List of Anatomy Level tasks.}
\label{anatomy_level_qa}
\vspace{-5mm}
\end{figure*}

\begin{figure*}[h]
\centering
\includegraphics[width=1\textwidth]{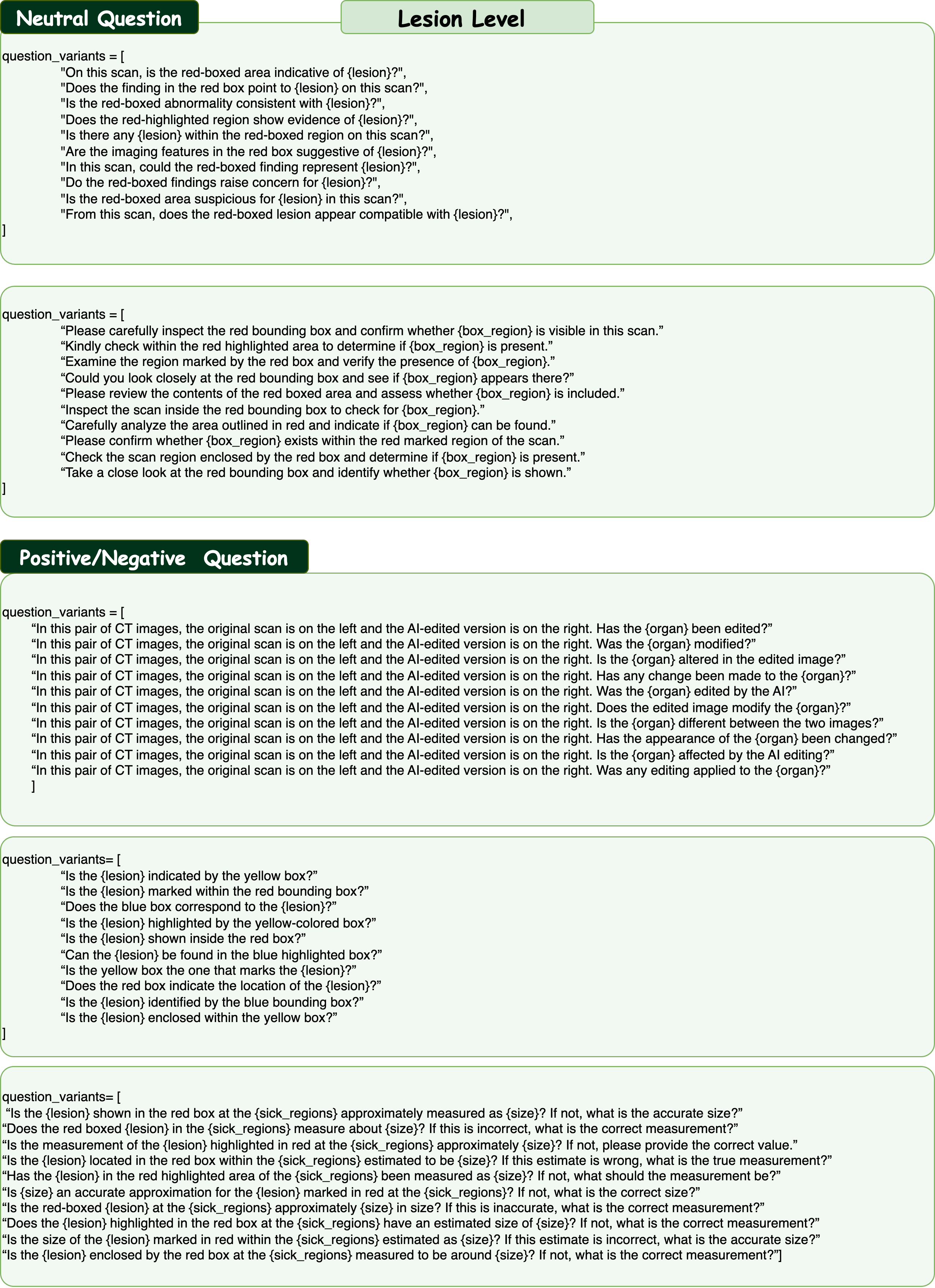} 
\vspace{-5mm}
\caption{Question List of Lesion Level tasks.}
\label{Lesion_level_qa}
\vspace{-5mm}
\end{figure*}

\end{document}